\documentclass[10pt]{article} 
\usepackage[accepted]{tmlr}

\usepackage{amsmath,amsfonts,bm}

\def\eqref#1{equation~\ref{#1}}

\def\1{\bm{1}}

\DeclareMathAlphabet{\mathsfit}{\encodingdefault}{\sfdefault}{m}{sl}
\SetMathAlphabet{\mathsfit}{bold}{\encodingdefault}{\sfdefault}{bx}{n}

\usepackage{hyperref}
\usepackage{url}

\usepackage{graphicx}
\usepackage{multirow}
\newcommand{\algname}{{SRR}}
\usepackage{url}            
\usepackage{booktabs}       
\usepackage{amsfonts}       
\usepackage{nicefrac}       
\usepackage{microtype}      
\usepackage{xcolor}         
\usepackage{graphicx}
\usepackage{multirow}
\usepackage{wrapfig}
\usepackage{bbm}
\usepackage{amsmath}
\usepackage{subcaption}
\usepackage{enumitem} 

\usepackage{algorithm}
\usepackage{algpseudocode}

\title{Self-supervise, Refine, Repeat: \\Improving Unsupervised Anomaly Detection}

\author{\name Jinsung Yoon, Kihyuk Sohn, Chun-Liang Li, Sercan \"{O}. Arik, Chen-Yu Lee, Tomas Pfister  \email \{jinsungyoon, kihyuks, chunliang, soarik, chenyulee, tpfister\}@google.com \\
      \addr Google Cloud AI
    }

\def\month{08}  
\def\year{2022} 
\def\openreview{\url{https://openreview.net/forum?id=b3v1UrtF6G}} 

\begin{document}

\maketitle

\begin{abstract}
Anomaly detection (AD), separating anomalies from normal data, has many applications across domains, from security to healthcare.
While most previous works were shown to be effective for cases with fully or partially labeled data, that setting is in practice less common due to labeling being particularly tedious for this task. 
In this paper, we focus on fully \emph{unsupervised} AD, in which the entire training dataset, containing both normal and anomalous samples, is unlabeled. 
To tackle this problem effectively, we propose to improve the robustness of one-class classification trained on self-supervised representations using a data refinement process. 
Our proposed data refinement approach is based on an ensemble of one-class classifiers (OCCs), each of which is trained on a disjoint subset of training data. 
Representations learned by self-supervised learning on the refined data are iteratively updated as the data refinement improves. 
We demonstrate our method on various unsupervised AD tasks with image and tabular data.
With a 10\% anomaly ratio on CIFAR-10 image data / 2.5\% anomaly ratio on Thyroid tabular data, the proposed method outperforms the state-of-the-art one-class classifier by 6.3 AUC and 12.5 average precision / 22.9 F1-score.
\end{abstract}

\section{Introduction}
\label{sec:intro}
Anomaly detection (AD), the task of distinguishing anomalies from normal data, plays crucial role in many real-world applications such as detecting faulty products from vision sensors in manufacturing, fraudulent behaviors at credit card transactions, or adversarial outcomes at intensive care units such as death, heart attack, or blood poisoning.

%

\begin{figure*}[t]
    \centering
    \includegraphics[width=\textwidth]{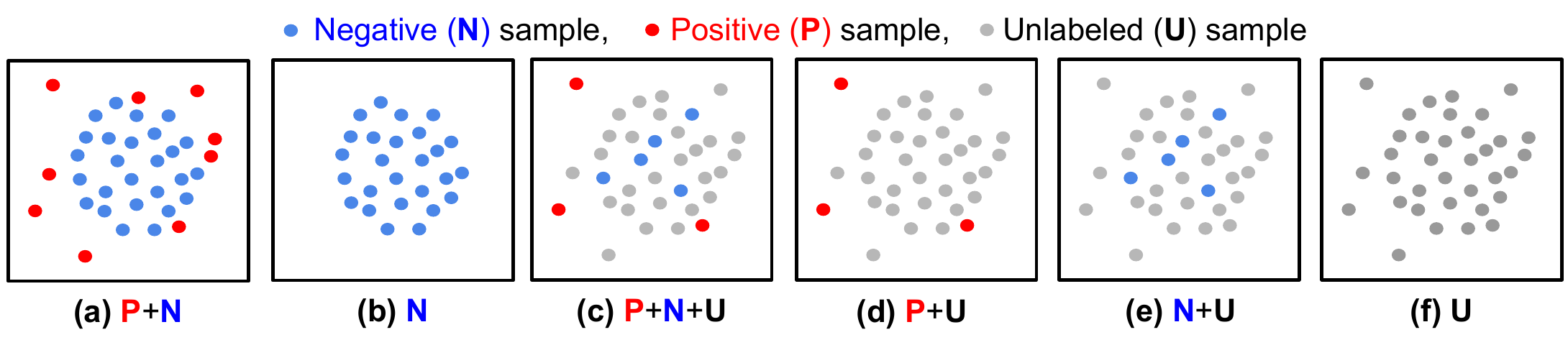}
    \caption{Anomaly Detection (AD) problem settings.
    Blue and red dots are for {\bf labeled} negative (normal) and positive (anomalous) samples, respectively. Grey dots denote {\bf unlabeled} samples. 
    While previous works mostly focus on supervised (a, b) or semi-supervised (c, d, e) settings, we tackle an AD problem using only unlabeled data (f) that may contain both negative and positive samples. 
    }
    \label{fig:general_problem_setting}
\end{figure*}

AD has been considered under various settings based on the availability of negative (normal) and positive (anomalous) data and their labels at training, as overviewed in Sec.~\ref{sec:related}. Each application scenario is dominated by different challenges.
When all positive and negative samples are available along with their labels (Fig.~\ref{fig:general_problem_setting}a), the problem can be treated as supervised classification and the dominant challenge becomes the imbalance in label distributions~\citep{chawla2002smote,estabrooks2004multiple,hwang2011new,barua2012mwmote,lee2000noisy,liu2007generative}. 
When only negative labeled data are available (Fig.~\ref{fig:general_problem_setting}b), the problem is `one-class classification' ~\citep{scholkopf1999support,tax2004support,ruff2018deep,hendrycks2018deep,golan2018deep,sohn2020learning,li2021cutpaste}. 
Various works have also extended approaches designed for these to settings with additional unlabeled data (Fig.~\ref{fig:general_problem_setting}c,d,e)~\citep{zhang2008learning,blanchard2010semi,gornitz2013toward,ruff2020deep} in a semi-supervised setting. 
While there exist many prior works in these settings, most of them rely on some labeled data, which is not desirable in many application scenarios. 

\begin{wrapfigure}{h!}{0.4\textwidth}
  \begin{center}
    \includegraphics[width=0.4\textwidth]{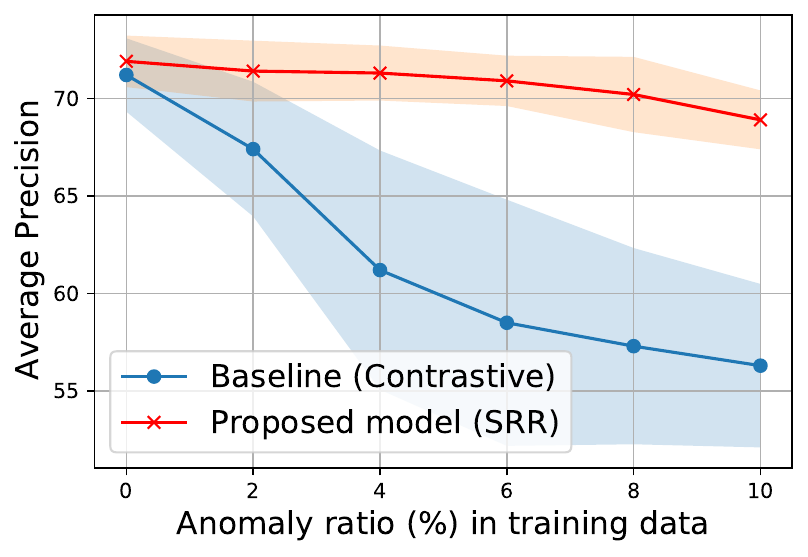}
  \end{center}
  \caption{Performance of the proposed model ({\algname}) and a baseline OCC using contrastive representation~\citep{sohn2020learning} on CIFAR-10 with different anomaly ratios in the training data.}
  \label{fig:teaser_perf}
\end{wrapfigure}

Unsupervised AD, on the other hand, poses unique challenges in the absence of any labeled data information, and a straightforward adaption of methods developed with the assumption of labeled data would be suboptimal. 
For example, some recent studies~\citep{zong2018deep,bergman2020classification} have applied one-class classifiers (OCCs) that are known to yield the state-of-the-art performance when trained on negative samples~\citep{golan2018deep,hendrycks2018deep,bergman2020classification,sohn2020learning,li2021cutpaste} to unsupervised AD, but their performance for unsupervised AD (when the unlabeled data contain both positive and negative samples) has been significantly degraded.
Fig.~\ref{fig:teaser_perf} illustrates this, showing the unsupervised AD performance of state-of-the-art Deep OCCs~\citep{sohn2020learning} with different anomaly ratios in unlabeled training data -- the average precision significantly drops even with a small contamination ratio (2\%) of the training data. 

Our framework {\algname} (\textbf{S}elf-supervise, \textbf{R}efine, \textbf{R}epeat), overviewed in Fig.~\ref{fig:main_block_diagram}, brings a data-centric\footnote{Definitions of the data-centric approaches can be found here (\url{https://spectrum.ieee.org/andrew-ng-data-centric-ai}).} approach to unsupervised AD with the principles of self-supervised learning without labels and iterative data refinement based on the agreement of OCC outputs. 
We propose to improve the state-of-the-art performance of OCCs, e.g. ~\citep{sohn2020learning,li2021cutpaste}, by refining the unlabeled training data so as to address the challenges elaborated above.
{\algname} iteratively trains deep representations using refined data while improving the refinement of unlabeled data by excluding potentially-positive (anomalous) samples. 
For the data refinement process, we employ an ensemble of OCCs, each of which is trained on a disjoint subset of unlabeled training data. 
The samples are declared as normal if there is a consensus between all the OCCs. 
The refined training data are used to train the final OCC to generate the anomaly scores in the unsupervised setting. 

Most prior AD works~\citep{golan2018deep,hendrycks2018deep,sohn2020learning,li2021cutpaste} assume that the data contain entirely negative samples, which makes them not truly unsupervised as they require having humans to do the data filtering. Most prior works are not actually designed for fully-unsupervised anomaly detection.
Some prior unsupervised AD works~\citep{ruff2018deep,zong2018deep,bergman2020classification} consider evaluating on an unsupervised setting where there exist a small percentage of anomalous samples in the training data, i.e. operating in `truly' unsupervised setting without having the need for humans to do any filtering in the training data.
However, these methods often suffer from significant performance degradation as the ratio of anomalous sample ratio increases (see Sec.~\ref{sec:exp_image}). Our method distinguishes from the literature by bringing a data-centric approach to unsupervised AD beyond the model-centric approaches.
We show the value of {\algname} in improving robustness of the performance as the anomaly ratio increases, as shown in Fig.~\ref{fig:teaser_perf}.

We conduct extensive experiments across various datasets from different domains, including semantic AD (CIFAR-10~\citep{krizhevsky2009learning}, Dog-vs-Cat~\citep{elson2007asirra}), 
real-world manufacturing visual AD use case (MVTec~\citep{bergmann2019mvtec}), and real-world tabular AD benchmarks (e.g., detecting medical or network anomalies).
We consider methods with both shallow ~\citep{scholkopf1999support,liu2008isolation} and deep~\citep{sohn2020learning,li2021cutpaste,bergman2020classification} models. 
We evaluate models at different anomaly ratios of unlabeled training data and show that {\algname} significantly boosts performance. 
For example, in Fig.~\ref{fig:teaser_perf}, {\algname} improves more than 15.0 average precision (AP) with a 10\% anomaly ratio compared to a state-of-the-art one-class deep model~\citep{sohn2020learning} on CIFAR-10.
Similarly, on MVTec, {\algname} retains a strong performance, dropping less than 1.0 AUC with 10\% anomaly ratio, while the best existing OCC~\citep{li2021cutpaste} drops more than 6.0 AUC.
The contributions of this paper are summarized below.
\begin{itemize}[noitemsep,nolistsep,leftmargin=*]
    \item We propose a novel data-centric framework, {\algname}, for unsupervised anomaly detection using the ensemble of one-class classifiers as a data refinement module.
    \item The proposed framework is a model-agnostic approach that can be applicable on top of any anomaly detection framework, considering different ways of applying self-supervised representation learning or one-class classification. 
    \item {\algname} achieves significant robustness improvements with various anomaly ratios -- in other words, the users do not need to worry about manually filtering the possible anomalies from the training data to minimize contamination. We demonstrate the superior performances on multiple tabular and image datasets.
\end{itemize}

\section{Related Work}\label{sec:related}
There are various existing works under different settings described in Fig.~\ref{fig:general_problem_setting}:

\textbf{Learning from both positives \& negatives setting} is often considered as a supervised binary classification problem. 
The challenge arises due to the imbalance in label distributions as positive samples are rare. 
As summarized in~\citep{branco2015survey}, to address this, over-/under-sampling ~\citep{chawla2002smote,estabrooks2004multiple}, weighted optimization ~\citep{hwang2011new,barua2012mwmote}, synthesizing data of minority classes~\citep{lee2000noisy,liu2007generative}, and hybrid methods~\citep{galar2011review} have been studied.

\textbf{Learning only from the negatives setting} is often converted to a one-class classification (OCC) problem, with the goal of finding a decision boundary that includes as many one-class samples as possible. 
Shallow models for this setting include one-class support vector machines~\citep{scholkopf1999support} (OC-SVM), support vector data description~\citep{tax2004support} (SVDD), kernel density estimation (KDE)~\citep{latecki2007outlier}, and Gaussian density estimation (GDE)~\citep{reynolds2009gaussian}. 
There are also auto-encoder based models \citep{zhou2017anomaly} that treat the reconstruction error as the anomaly score. 
Deep learning based OCCs have been developed, such as Deep OCC~\citep{ruff2018deep}, geometric transformation~\citep{golan2018deep}, or outlier exposure~\citep{hendrycks2018deep}. 
Noting the degeneracy or inconsistency of learning objectives of existing end-to-end trainable Deep OCCs, \citep{sohn2020learning} propose a deep representation OCC, a two-stage framework that learns self-supervised representations~\citep{gidaris2018unsupervised,chen2020simple} followed by shallow OCCs. That work is extended for texture anomaly localization with CutPaste~\citep{li2021cutpaste}. 
Robustness against very low anomaly ratios of these in unsupervised setting is explored in~\citep{zong2018deep,bergman2020classification}.

\textbf{The semi-supervised learning setting} utilizes a small set of labeled samples and large set of unlabeled samples to distinguish anomalies from normal data. 
Depending on which labeled samples are given, this setting can be split into three sub-categories. 
When only some positive/negative labeled samples are provided, we denote that as a PU (positive + unlabeled) / NU (negative + unlabeled) setting. 
Most previous works in semi-supervised AD settings focus on the NU setting where only some of the normal labeled samples are given \citep{munoz2010semisupervised,song2017hybrid,akcay2018ganomaly}. 
The PNU (positive + negative + unlabeled) setting is the more general semi-supervised setting where subsets of both positive and negative labeled samples are given. 
Deep SAD \citep{ruff2020deep} and SU-IDS \citep{min2018ids} belong to this category. We show significant outperformance of {\algname} compared to Deep SAD, on multiple benchmark datasets without using any labeled data (see Sec. \ref{sec:exp_image}).


\textbf{The unlabeled setting} has received relatively less attention despite its significance in minimizing labeling costs. 
The popular methods for this setting include isolation forest~\citep{liu2008isolation} and local outlier factor~\citep{breunig2000lof}. These are difficult to scale, and less compatible with recent advances in representation learning. 
While OCCs, such as OC-SVM, SVDD, or their deep counterparts, can also be applied to unlabeled settings by assuming the data are all negative, and the robustness of those methods has been demonstrated in part~\citep{zong2018deep,bergman2020classification}. In practice, we observe a significant performance drop with a high anomaly ratio, shown in Fig.~\ref{fig:teaser_perf}.  
In contrast, our proposed framework is able to maintain high performance across different anomaly ratios.

\textbf{Data refinement} has been applied to AD in some prior work. \citep{pang2020self,beggel2019robust} generate pseudo-labels using binary classification and OC-SVM for data refinement to boost the consequent AD performance in unsupervised settings. \citep{mohseni2021can,xia2015learning,zhou2017anomaly,lai2019robust,berg2019unsupervised} use the reconstruction errors of the auto-encoder as an indicator for removing possible anomalies. 
\citep{feng2021mist} and \citep{meng2021semi} use data refinement for AD in supervised and semi-supervised settings. 
{\algname} differentiates from \citep{xia2015learning,beggel2019robust,pang2020self} in multiple key ways:
\begin{itemize}[noitemsep,nolistsep,leftmargin=*]
\item \citep{xia2015learning,beggel2019robust} are based on auto-encoders and they directly utilize the reconstruction errors as informative signals for AD and iterative data refinement. However, prior work \citep{ren2019likelihood} discovered that the reconstruction is not a good informative signal for outlier detection. Also as shown in Fig.~\ref{fig:main_cifar_result}, the performance of DAE (reconstruction errors based AD) is much worse than alternatives. 
\item {\algname} utilizes ensemble learning for data refinement to improve the robustness which is critical in unsupervised AD. On the other hand, \citep{xia2015learning,beggel2019robust,pang2020self} utilize a single model for data refinement. As can be seen in Fig.~\ref{fig:ablation_result_k}, the impact of the ensemble model for data refinement is significant. 
\item \citep{xia2015learning,beggel2019robust,pang2020self} directly utilize the (pseudo) abnormal samples for model training in addition to (pseudo) normal samples. For instance, the model in \citep{pang2020self} is trained via two-class ordinal regression where two classes come from (pseudo) normal and (pseudo) abnormal samples. Directly relying on the (imperfectly) labeled abnormal samples can be harmful for AD due to the overfitting problems and high False Positive Rates (FPR) in (pseudo) abnormal samples.\footnote{For instance, with 80\% of the recall (for the abnormal sample discovery), the precision (for the abnormal sample discovery) is 28.2\% for CIFAR-10 and 31.7\% with MVTec (with 6\% abnormal ratio) using {\algname} framework -- the majority of the (pseudo) abnormal samples are actually the normal samples (71.8\% for CIFAR-10 and 68.3\% for MVTec datasets). This would be a strong evidence that directly utilizing the (pseudo) abnormal samples for training can be harmful.}
\item \citep{xia2015learning,pang2020self} are based on the end-to-end AD models which are empirically less accurate than two-stage models \citep{sohn2020learning}. 
\end{itemize}

\textbf{Self-training}~\citep{scudder1965probability,mclachlan1975iterative} is an iterative training mechanism using predicted pseudo labels as targets. 
It has regained popularity recently with its successful results in semi-supervised image classification~\citep{berthelot2019mixmatch,sohn2020fixmatch,xie2020self}. 
To improve the quality of pseudo labels, employment of an ensemble of classifiers has also been studied. \citep{brodley1996identifying} trains an ensemble of classifiers with different classification methods to make a consensus for noisy label verification, while co-training~\citep{blum1998combining} trains multiple classifiers, each of which is trained on the distinct views, to supervise other classifiers. 
Co-teaching~\citep{han2018co} and DivideMix~\citep{li2019dividemix} share similar ideas -- they both train multiple deep neural networks on separate data batches to learn different decision boundaries, thus, becoming useful for noisy label verification. 
While sharing a similarity, {\algname} has clear differences from the previous works -- {\algname} performs \emph{iterative training} with data refinement (with robust \emph{ensemble} methods) and self-supervised learning for \emph{unsupervised} AD.

%

\section{Proposed Framework}
\label{sec:method}

\begin{figure*}[t!]
    \centering
    \includegraphics[width=0.9\textwidth]{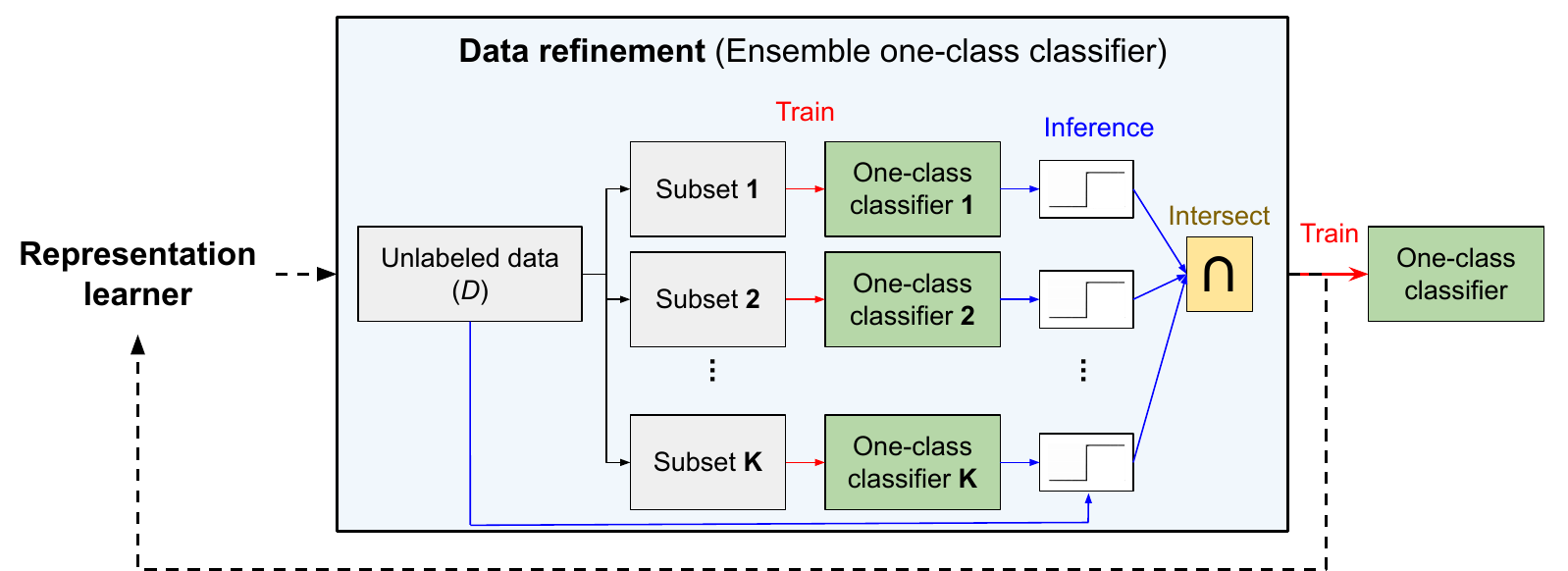}
    \caption{Block diagram of {\algname} composed of representation learner (Sec.~\ref{sec:method_model_update}), data refinement (Sec.~\ref{sec:method_data_refinement}), and final OCC blocks.
    The representation learner updates the corresponding deep neural network models using refined data from the data refinement block. 
    Data refinement is implemented by an ensemble of OCCs, each of which is trained on $K$ disjoint subsets of unlabeled training data. 
    Samples predicted as normal by all classifiers are retained in the refined data, and are used to update the representation learner and final OCC.
    The training process of {\algname} is repeated until the convergence of the representation learner's loss (if no improvement is observed for 5 epochs). Convergence graphs are provided in the Appendix~\ref{sect:appendix_convergence}.
    %
    }
    \label{fig:main_block_diagram}
\end{figure*}

%
\textbf{S}elf-supervise, \textbf{R}efine, and \textbf{R}epeat ({\algname}) is an iterative training framework, where we refine the data (Sec.~\ref{sec:method_data_refinement}) and update the representation with the refined data (Sec.~\ref{sec:method_model_update}), followed by OCC training on refined representations. 
Fig.~\ref{fig:main_block_diagram}  overviews the framework and Algorithm~\ref{alg:main_pseudocode} provides the pseudo-code. 
%


\textbf{Notation.} 
We denote the training data as $\mathcal{D} \,{=}\, \{\textbf{x}_i\}_{i=1}^N$ where $\textbf{x}_i \,{\in}\, \mathcal{X}$ and $N$ is the number of training samples. $y_i \,{\in}\, \{0, 1\}$ is the corresponding label to $\textbf{x}_i$, where $0$ denotes normal (negative) and $1$ denotes anomaly (positive). Note that labels are not provided in the unsupervised setting. 

Let us denote a feature extractor as $g: \mathcal{X} \rightarrow \mathcal{Z}$. $g$ may include any data preprocessing functions, an identity function (if raw data is directly used for one-class classification), and learned or learnable representation extractors such as deep neural networks.
%
%
Let us define an OCC as $f: \mathcal{Z} \rightarrow [-\infty, \infty]$ that outputs anomaly scores given the input features $g(\textbf{x})$, such that the higher the score $f(g(\textbf{x}))$ is, the more anomalous the sample $\mathbf{x}$ would be predicted as.
The binary AD is made after thresholding: $\mathbbm{1}\big(f(g(\textbf{x}))\geq\eta\big)$.

\subsection{Data refinement}
\label{sec:method_data_refinement}

A naive way to generate pseudo labels for unlabeled data would be to construct an OCC on raw data or learned representations as in \citep{sohn2020learning}, pick a threshold for the anomaly score and obtain a binary label for samples as normal vs. anomalous. 
As we update the model with refined data that exclude samples that are predicted to be anomalous, it is important to generate pseudo labels for training data accurately. 
To this end, instead of training a single classifier (unlike most previous works on data refinement for AD \citep{pang2020self,beggel2019robust,zhou2017anomaly}), we train an ensemble of $K$ OCCs and aggregate their predictions to generate pseudo labels. 

We illustrate the data refinement block in Fig.~\ref{fig:main_block_diagram} and as \textproc{RefineData} in Algorithm~\ref{alg:main_pseudocode}. 
Specifically, we randomly divide the unlabeled training data $\mathcal{D}$ into $K$ disjoint subsets $\mathcal{D}_{1},...,\mathcal{D}_{K}$, and train $K$ different OCCs $(f_{1},...,f_{K})$ on corresponding subsets $(\mathcal{D}_{1},...,\mathcal{D}_{K})$. Then, we estimate a binary pseudo-label of the data $\mathbf{x}_{i}\,{\in}\,\mathcal{D}$ as follows:
\begin{align}
    \hat{y}_{i} & = 1 - \prod_{k=1}^{K}\Big[1-\mathbbm{1}\big(f_{k}(g(\mathbf{x}_{i})) \geq \eta_{k}\big)\Big] \label{eq:prediction_aggregation} \\
    \eta_{k} &= \max \eta \,\mbox{ s.t. }\,  \frac{100}{N}\sum_{i=1}^N \mathbbm{1}\big(f_k(g(\textbf{x}_i)) \geq \eta\big) \geq \gamma \label{eq:threshold}
\end{align}
where $\mathbbm{1}(\cdot)$ is the indicator function that outputs 1/0 if the input is True/False. $f_{k}(g(\mathbf{x}_{i}))$ represents an anomaly score of $\mathbf{x}_{i}$ for an OCC $f_{k}$. 
$\eta_{k}$ in Eq.~\ref{eq:threshold} is a threshold determined as a $\gamma$ percentile of the anomaly score distribution $\{f_{k}(g(\mathbf{x}_{i}))\}_{i=1}^{N}$.

%
To interpret Eq.~\ref{eq:prediction_aggregation}, $\mathbf{x}_{i}$ is predicted as normal, i.e. $\hat{y}_{i}\,{=}\,0$, if all $K$ OCCs predict it as normal. 
While this may be too strict and potentially reject many true normal samples in the training set, we find that empirically, it is critical to be able to exclude true anomalous samples from the training set. 
The effectiveness of employment of an ensemble of OCCs  to improve the robustness against overfitting of OCCs is empirically shown in Sec.~\ref{sec:exp_ablation}. More specifically, Fig~\ref{fig:ablation_result_k} shows that the performance is much better for higher ensemble counts, compared to a single classifier training along with the feature extractor.

\subsection{Representation update}
\label{sec:method_model_update}

{\algname} follows the idea of deep representation learning to perform OCCs on~\citep{sohn2020learning}, where in the first stage, a deep neural network is trained with self-supervised learning (such as rotation prediction~\citep{golan2018deep}, contrastive~\citep{sohn2020learning}, or CutPaste~\citep{li2021cutpaste}) to obtain meaningful representations of the data, and in the second stage OCCs are trained on these learned representations. 
Such a two-stage framework is shown to be beneficial as it prevents the `hypersphere collapse' of the deep OCCs by the favorable inductive bias it brings with the architectural constraints ~\citep{ruff2018deep}. 

We propose to conduct self-supervised representation learning jointly with data refinement. 
More precisely, we train a feature extractor $g$ using $\hat{\mathcal{D}}\,{=}\,\{\mathbf{x}_{i}\,|\,\hat{y}_{i}\,{=}\,0\}$, a subset of unlabeled data $\mathcal{D}$ that only includes samples whose predicted labels with an ensemble OCC from Sec.~\ref{sec:method_data_refinement} are negative. We also update $\hat{\mathcal{D}}$ as we proceed with representation learning. Overall method is illustrated in Algorithm~\ref{alg:main_pseudocode}.
Unlike previous works~\citep{sohn2020learning,li2021cutpaste} that use the entire training data for representation learning, we empirically show that it is beneficial to refine the training data even for learning of the deep representations. 
Without representation refinement, the performance improvements of {\algname} are limited, as shown in Sec.~\ref{sec:exp_ablation}.
%
Lastly, to get predictions at test-time, we train an OCC on refined data $\hat{\mathcal{D}}$ on the updated representations by $g$ as in line 13-14 in Algorithm~\ref{alg:main_pseudocode}.


\begin{algorithm}[t]
    \caption{{\algname}: Self-supervise, Refine, Repeat.}
    \label{alg:main_pseudocode}
    \begin{flushleft}
        \textbf{Input}: Train data $\mathcal{D} = \{\textbf{x}_i\}_{i=1}^N$, Ensemble count ($K$), threshold ($\gamma$)\\
        \textbf{Output}: Refined data ($\hat{\mathcal{D}}$), trained OCC ($f$), feature extractor ($g$)\\
    \end{flushleft}
    \begin{algorithmic}[1]
        \Function {RefineData}{$\mathcal{D},g,K,\gamma$}
        \State Train OCC models $\{f_k\}_{k=1}^{K}$ on $\{\mathcal{D}_k\}_{k=1}^{K}$, $K$ disjoint subsets of the training data $\mathcal{D}$.
        \State Compute thresholds $\eta_{k}$'s for $\gamma$ percentile of anomaly distributions using Eq.~\ref{eq:threshold}.
        \State Predict binary labels $\hat{y}_{i}$ using Eq.~\ref{eq:prediction_aggregation}.
        \State Return $\hat{\mathcal{D}} = \{\mathbf{x}_{i}:\hat{y}_{i}=0, \mathbf{x}_{i}\,{\in}\,\mathcal{D}\}$.
        \EndFunction
    \Function {\algname}{$\mathcal{D},K,\gamma$}
    \State Initialize the feature extractor $g$.
    \While{$g$ not converged}
        \State $\hat{\mathcal{D}}$ = \textproc{RefineData}($\mathcal{D},g,K,\gamma$).
        \State Update $g$ using $\hat{\mathcal{D}}$ with self-supervised objectives.
    \EndWhile
    \State $\hat{\mathcal{D}}$ = \textproc{RefineData}($\mathcal{D},g,K,\gamma$).
    \State Train an OCC model ($f$) on refined data $(\hat{\mathcal{D}})$.
    \EndFunction
    \end{algorithmic}
\end{algorithm}

\subsection{Unsupervised model selection}
\label{sec:method_data_refinement_params}

As {\algname} is designed for unsupervised AD, labeled validation data for hyperparameter tuning are typically not available and \textit{the framework should enable robust model selection without any reliance on labeled data}. Here, we provide insights on how to select important hyperparameters, and later in Sec.~\ref{sec:exp_ablation_sensitivity}, we perform sensitivity analyses for these hyperparameters.

Data refinement of {\algname} introduces two hyperparameters: the number of OCCs ($K$) and the percentile threshold ($\gamma$). 
There is a trade-off between the number of classifiers for the ensemble and the size of disjoint subsets for training each classifier. 
With large $K$, we aggregate prediction from many classifiers, each of which may contain randomness from training. 
This comes at the cost of reduced performance per classifier as we use smaller subsets to train them. 
In practice, we find $K\,{=}\,5$ works well across different datasets and anomaly ratios.
$\gamma$ controls the purity and coverage of refined data -- if $\gamma$ is large, and thus classifiers reject too many samples, the refined data could be more pure and contain mostly the normal samples; however, the coverage of the normal samples would be limited. 
On the other hand, with a small $\gamma$, the refined data may still contain many anomalies and the performance improvement with {\algname} would be limited. 
We empirically observe that {\algname} is robust to the selection of $\gamma$ when it is chosen from a reasonable range. 
In our experiments, we find 1-2 times of the true anomaly ratio to be often a reasonable choice. 
In other words, it is safer to use $\gamma$ higher than the expected true anomaly ratio. 
In some cases, the true anomaly ratio may not be available. For such scenarios, we propose Otsu's method \citep{sezgin2004survey} to estimate the anomaly ratio of the training data for determining the threshold $\gamma$ (experiment results are in Sec.~\ref{sect:otsu_method}).



\section{Anomaly Detection Performance}\label{sec:experiment}
We evaluate the efficacy of our proposed framework for unsupervised AD tasks for tabular (Sec.~\ref{sec:exp_tabular}) and image (Sec.~\ref{sec:exp_image}) data types. 
We experiment varying ratios of anomaly samples in unlabeled training data and with different combinations of representation learning and OCCs. 
In Sec.~\ref{sec:exp_ablation}, we provide performance analyses to better explain major constituents of the performance, as well as the sensitivity analyses.


\textbf{Implementation details:} 
To reduce the computational complexity of the data refinement block, we utilize a simple OCC such as GDE in the data refinement block. 
In a two-stage model, we only update the data refinement block at 1st, 2nd, 5th, 10th, 20th, 50th, 100th, 500th epochs. 
After 500 epochs, we update the data refinement block per each 500th epoch. 
Each experimental run is performed on a single V100 GPU.  Additional discussions can be found in Appendix~\ref{sect:appendix_computational_complexity}.



\subsection{Experiments on tabular data}
\label{sec:exp_tabular}
\textbf{Datasets.}
Following~\citep{zong2018deep,bergman2020classification}, we test the performance of {\algname} on a variety of real-world tabular AD datasets, including network (KDDCup) and medical (Thyroid, Arrhythmia) AD from the UCI repository~\citep{asuncion2007uci}. 
We also use KDDCup-Rev, where the labels of KDDCup are reversed so that an attack represents the anomaly~\citep{zong2018deep}. 
To construct the data splits, we utilize 50\% of normal samples for training. 
In addition, we hold out some anomaly samples (amounting to 10\% of the normal samples) from the data. 
This allows to simulate unsupervised settings with an anomaly ratio of up to 10\% of entire training set. 
Rest of the data is used for testing.\footnote{Note that the experimental settings with contaminated training data in GOAD~\citep{bergman2020classification} and DAGMM~\citep{zong2018deep} are slightly different from ours. 
The contamination ratio in this paper is defined as the anomaly ratio over the entire training data, while their contamination ratio is defined as the anomaly ratio over all the anomalies in the dataset.}
We conduct experiments using 5 random splits and 5 random seeds, and report the average and standard deviation of 25 F1-scores (with scale 0-100) for the performance metric.



\textbf{Models.}
We focus on comparisons with GOAD \citep{bergman2020classification} (the state-of-the-art AD model in the tabular domain), with and without {\algname}.
GOAD utilizes random transformation classification as the pretext task of self-supervised representation learning, and the normality score is determined by whether transformations are accurately included in the transformed space of the normal samples. 
We re-implement GOAD~\citep{bergman2020classification}
with a few modifications. 
First, instead of using embeddings to compute the loss, we use a parametric classifier, similarly to augmentation prediction~\citep{sohn2020learning}. 
Second, we follow the two-stage framework~\citep{sohn2020learning} to construct deep OCCs. 
For the clean training data setting, our implementation achieves 98.0 for KDD, 95.0 for KDD-Rev, 75.1 for Thyroid, and 54.8 for Arrhythmia F1-scores, which are comparable to those reported in \citep{bergman2020classification}. 
Please see Appendix~\ref{sec:app_goad} for formulation and implementation details.

We also include the comparisons with OC-SVM, with an radial basis function (rbf) kernel. 
For comparisons with other conventional baselines including Standard PCA \citep{wold1987principal}, Robust PCA \citep{candes2011robust}, and Local Outlier Factor \citep{breunig2000lof}, please see the Appendix (we do not present them here as they are far worse than the state-of-the-art on the experimented datasets).

\textbf{Results.}

\begin{figure*}[h!]
\centering
\begin{subfigure}[b]{0.24\textwidth}
  \includegraphics[width=\linewidth]{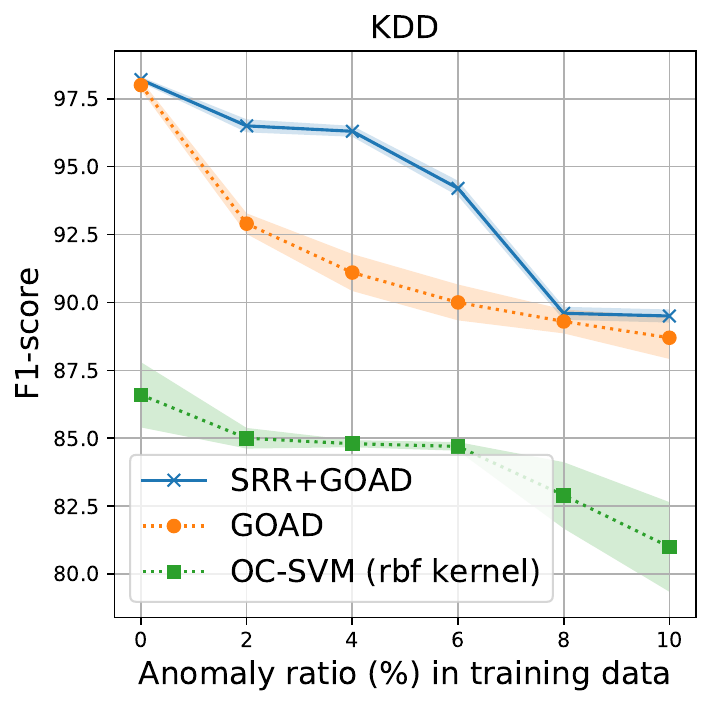}
\end{subfigure}
\begin{subfigure}[b]{0.24\textwidth}
  \includegraphics[width=\linewidth]{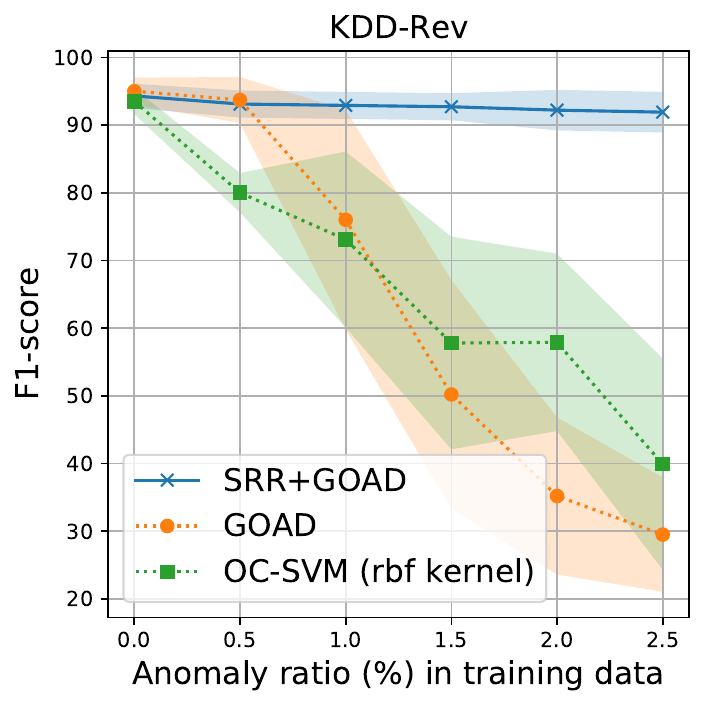}
\end{subfigure}
\begin{subfigure}[b]{0.24\textwidth}
  \includegraphics[width=\linewidth]{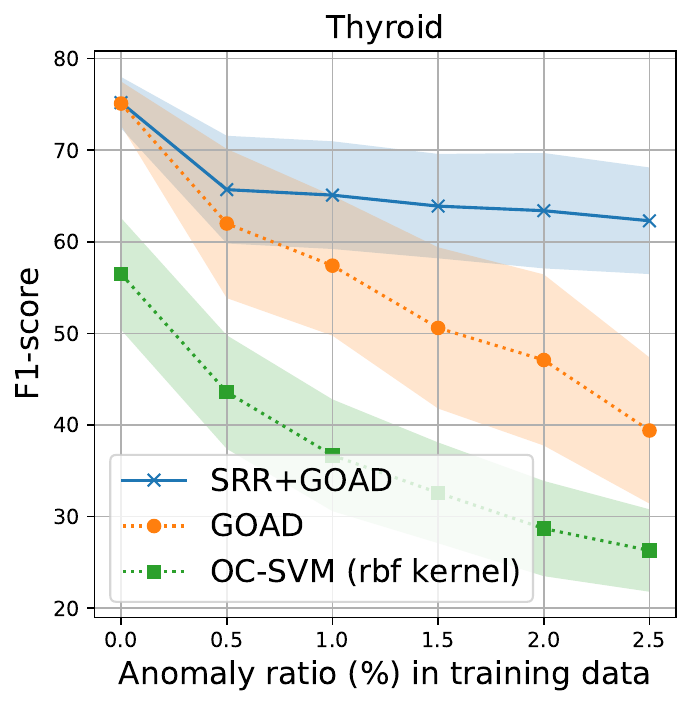}
\end{subfigure}
\begin{subfigure}[b]{0.24\textwidth}
  \includegraphics[width=\linewidth]{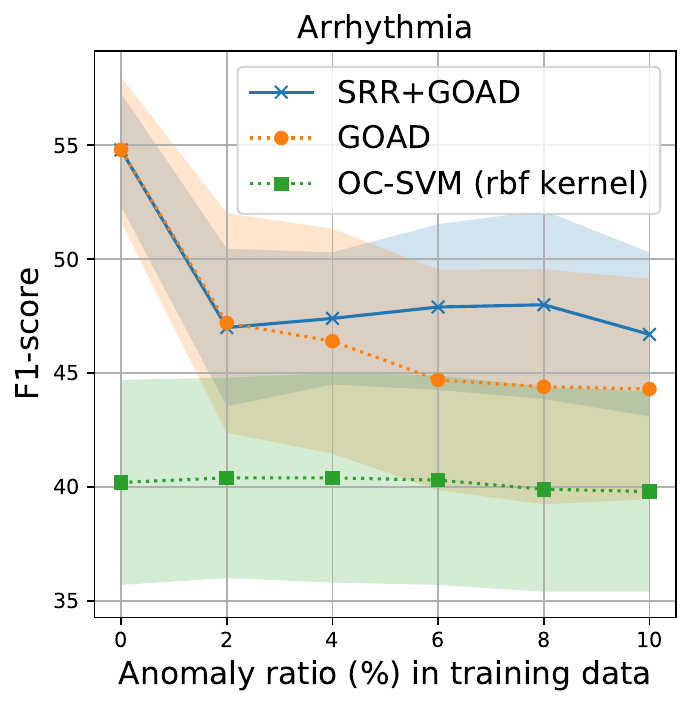}
\end{subfigure}
\caption{Unsupervised AD performance (F1-score) using OC-SVM (with rbf kernel), GOAD \citep{bergman2020classification}, and GOAD with the proposed method ({\algname}) on four tabular datasets. 
Shaded areas represent the standard deviations. 
}\label{fig:main_tabular_result}
\end{figure*}

Fig.~\ref{fig:main_tabular_result} presents the results for {\algname} compared to the GOAD baseline, as well as OC-SVM. 
Overall, we observe superior AD performance with {\algname} consistently across datasets. 
For KDD-Rev, the performance of GOAD (without {\algname}) drops significantly even with a small anomaly ratio in training data, and beyond 1\%, the drop becomes very significant. On the other hand, {\algname} can preserve the high anomaly detection performance despite the contamination in the training data, and at 2.5\% anomaly ratio, we can observe almost the double F-score. 
The improvements on other datasets are smaller, but similarly we can observe significant F-score differences when the training data is contaminated with many anomaly samples. On two small-scale datasets, Thyroid \& Arrhythmia, we observe more variance in the results due to data randomness when fitting the models, as expected. The improvements of {\algname} also highlights its data efficiency, as the proposed refinement mechanism can be effective even with a small subset of samples.

\subsection{Experiments on image data}
\label{sec:exp_image}

\textbf{Datasets.}
We evaluate {\algname} on various visual AD benchmarks, including real-world manufacturing AD dataset (MVTec~\citep{bergmann2019mvtec}) and semantic AD datasets (CIFAR-10~\citep{krizhevsky2009learning}, f-MNIST~\citep{xiao2017fashion}, Dog-vs-Cat~\citep{elson2007asirra}). 
For CIFAR-10, f-MNIST, and Dog-vs-Cat datasets, samples from one class are set to be normal and the rest from other classes are set to be abnormal.
Similar to the experiments on tabular data in Sec.~\ref{sec:exp_tabular}, we swap a certain amount of the normal training data with anomalies given the target anomaly ratio.
For MVTec, since there are no anomalous data available for training, we borrow 10\% of the anomalies from the test set and swap them with normal samples in the training set. 
Note that 10\% of samples borrowed from the test set are excluded from evaluation.
For all datasets, we experiment with varying anomaly ratios from 0\% to 10\%. 



We use AUC and average precision (AP) metrics to quantify the AD performance (with scale 0-100). 
When computing AP, we set the minority class of the test set as label 1 and majority as label 0. 
We run all experiments with 5 random seeds and report the average performance for each dataset across all classes.

%


\textbf{Models.}
For semantic AD benchmarks, CIFAR-10, f-MNIST, and Dog-vs-Cat, we compare the {\algname} with two-stage OCCs~\citep{sohn2020learning} using various representation learning methods, such as distribution-augmented contrastive learning~\citep{sohn2020learning}, rotation prediction~\citep{gidaris2018unsupervised,golan2018deep} and its improved version, and denoising autoencoder. 
For MVTec benchmarks, we use CutPaste~\citep{li2021cutpaste} as the baseline and compare to its version with {\algname} integration.
For both experiments, we use the ResNet-18 architecture, trained from random initialization, using the hyperparameters from \citep{sohn2020learning} and \citep{li2021cutpaste}. 
The same model and hyperparameter configurations are used for {\algname} with $K\,{=}\,5$ classifiers in the ensemble. 
We set $\gamma$ as twice the anomaly ratio of training data. 
For 0\% anomaly ratio, we set $\gamma$ as 0.5.
Finally, a Gaussian Density Estimator (GDE) on learned representations is used as the OCC.

For comparisons with other conventional baselines including Standard PCA \citep{wold1987principal}, Robust PCA \citep{candes2011robust}, and Robust autoencoder \citep{zhou2017anomaly}, please see the Appendix (we do not present them here as they are far worse than the state-of-the-art on the experimented datasets).



\textbf{Results.}

\begin{figure*}[h!]
\minipage{0.32\textwidth}
  \includegraphics[width=\linewidth]{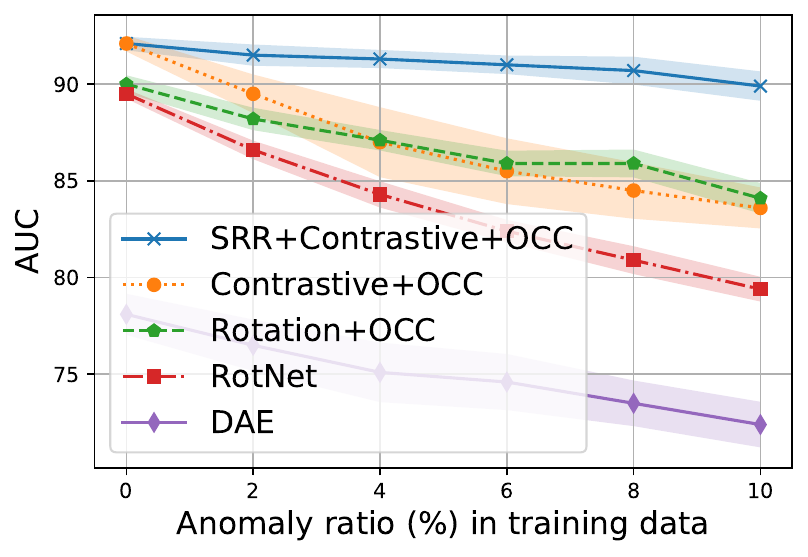}
\endminipage\hfill
\minipage{0.32\textwidth}
  \includegraphics[width=\linewidth]{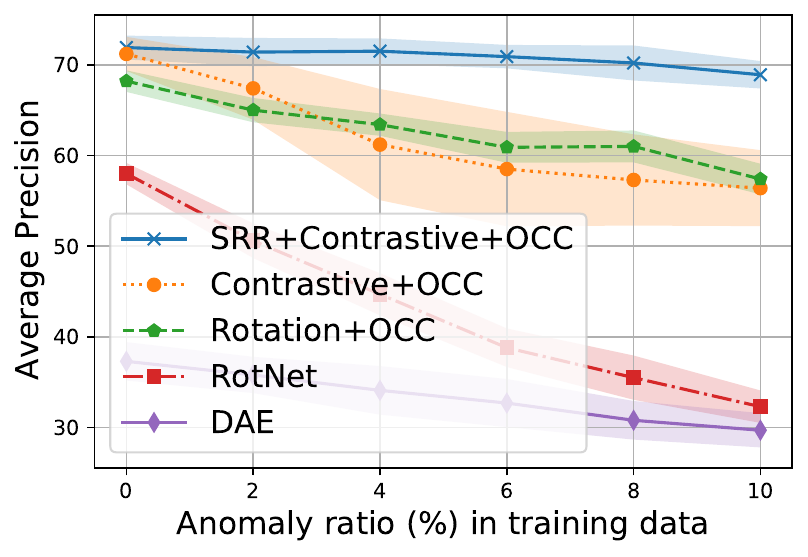}
\endminipage\hfill
\minipage{0.32\textwidth}
  \includegraphics[width=\linewidth]{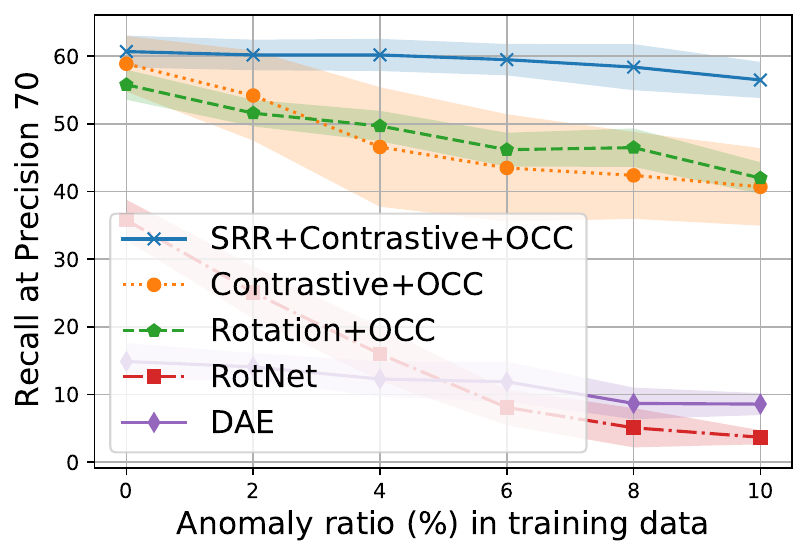}
\endminipage\hfill
\caption{Unsupervised AD performance with various OCCs on CIFAR-10 dataset. For {\algname} we adapt distribution-augmented contrastive representation learning \citep{sohn2020learning}. (Left) AUC, (Middle) Average Precision (AP), (Right) Recall at Precision 70.}\label{fig:main_cifar_result}
\end{figure*}

On CIFAR-10 dataset, Fig.~\ref{fig:main_cifar_result} compares {\algname} to other alternatives.
In general, most methods suffer from a significant performance drop with increased anomaly ratio, despite using representation learning.
For example, the AUC for AD based on distribution-augmented contrastive representation~\citep{sohn2020learning} drops from 92.1 to 83.6 when anomaly ratio becomes 10\%. 
Similarly, with the improved rotation prediction representation~\citep{gidaris2018unsupervised}, the AD performance drops from 90.0 to 84.1 in AUC.
On the other hand, {\algname} effectively handles the contamination in training data and the reduction in performance is much lower with increased anomaly ratio -- it achieves 89.9 AUC with 10\% anomaly ratio, reducing the performance drop by 74.1\% compared to the best alternative.
One `oracle' upper bound would be removal of all anomalies from training data, which is the same as the performance at 0\% anomaly ratio for the same size of the data. As Fig.~\ref{fig:main_cifar_result} shows, the performance of {\algname} is similar to this oracle upper bound, with less than 2.5 AUC difference, up until high anomaly ratios (10\%).
The results are also similar in other metrics, such as AP and Recall at Precision of 70, described in Fig.~\ref{fig:main_cifar_result}.


\begin{figure*}[h!]
\begin{subfigure}[b]{0.32\textwidth}
  \centering
  \includegraphics[width=\linewidth]{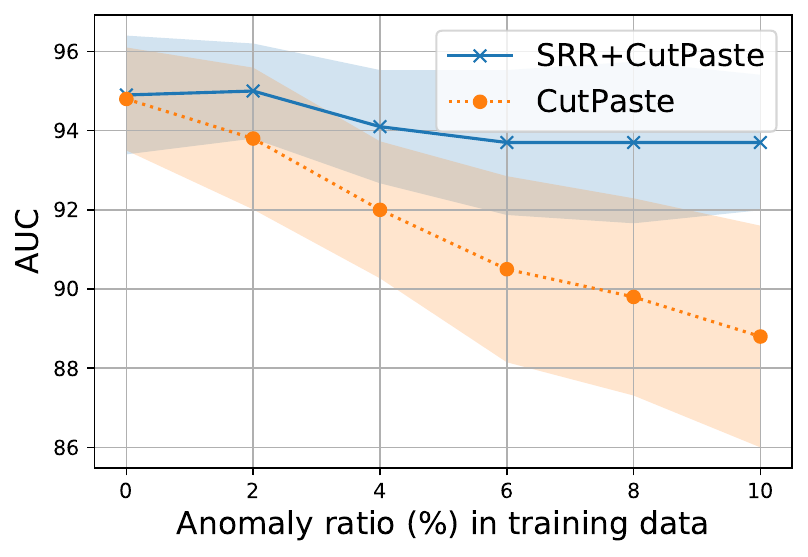}
  \includegraphics[width=\linewidth]{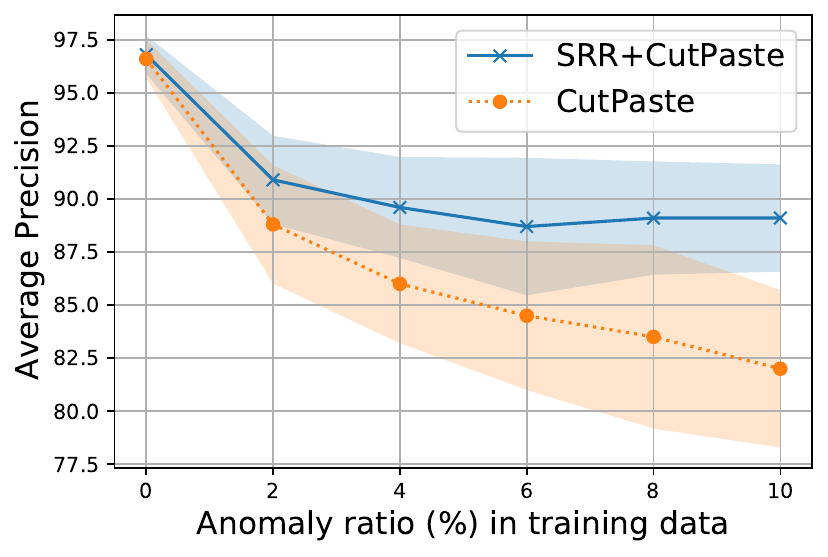}
  \caption{MVTec}
  \label{fig:image_result_mvtec}
\end{subfigure}
\begin{subfigure}[b]{0.32\textwidth}
  \includegraphics[width=\linewidth]{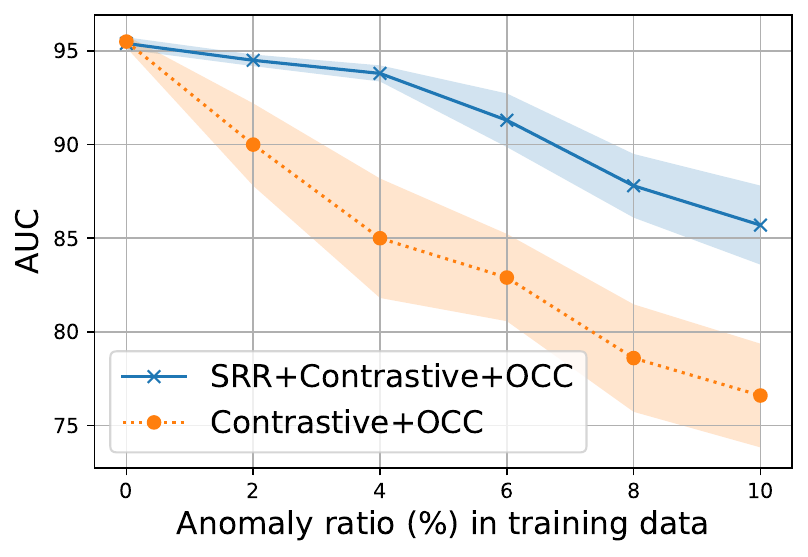}
  \includegraphics[width=\linewidth]{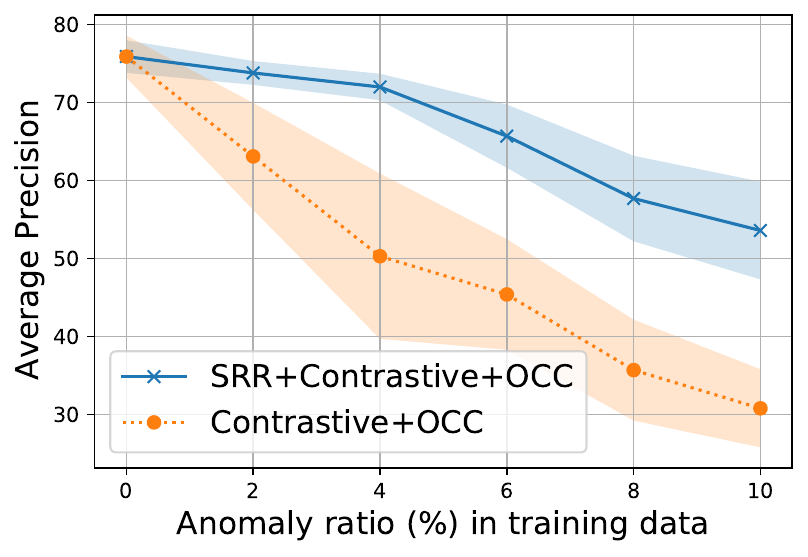}
  \caption{f-MNIST}
  \label{fig:image_result_fmnist}
\end{subfigure}
\begin{subfigure}[b]{0.32\textwidth}
  \includegraphics[width=\linewidth]{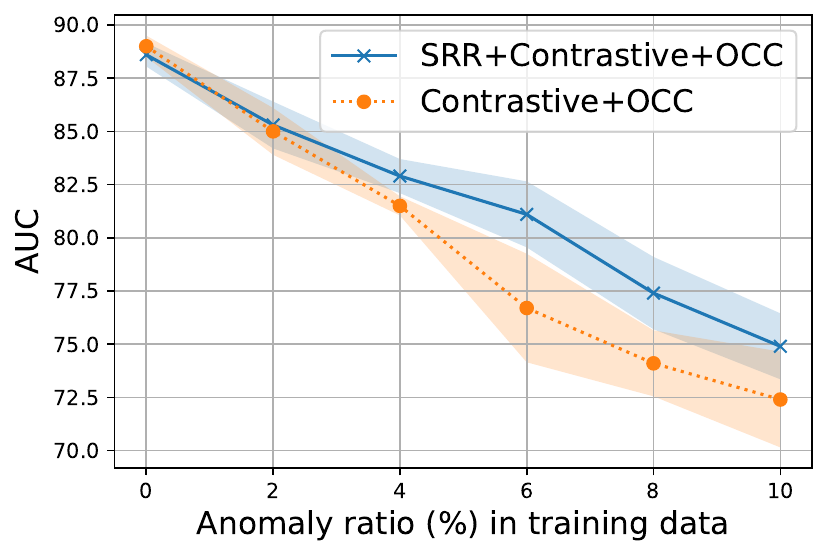}
  \includegraphics[width=\linewidth]{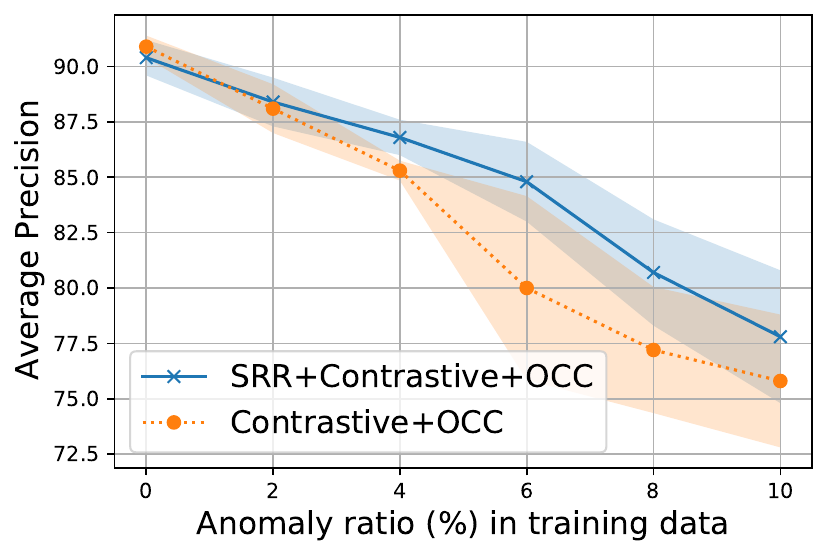}
  \caption{Dog-vs-Cat}
  \label{fig:image_result_dog_vs_cat}
\end{subfigure}
\caption{Unsupervised AD performance on (a) MVTec (b) f-MNIST, and (c) Dog-vs-Cat  datasets with varying anomaly ratios. 
We use state-of-the-art one-class classification models for baselines, such as distribution-augmented contrastive representations~\citep{sohn2020learning} for f-MNIST and Dog-vs-Cat, or CutPaste~\citep{li2021cutpaste} for MVTec, and build {\algname} on top of them.}
\label{fig:main_image_result}
\end{figure*}

Next, we show additional results on 3 visual AD datasets in Fig.~\ref{fig:main_image_result}, comparing to the best alternatives with representation learning and OCC. 
The improvements of {\algname}, particularly at higher anomaly ratios, are consistent across different cases. For example, on MVTec dataset, {\algname} improves AUC by 4.9 and AP by 7.1 compared to the state-of-the-art CutPaste OCC with an anomaly ratio of 10\%. On Dog-vs-Cat task, we observe the improvement to be smaller, which we attribute to similarity of the learned representation for these two classes, that makes anomaly definition more ambiguous.

\section{Performance Analyses}
\label{sec:exp_ablation}

In this section, we focus on providing more insights on the important constituents of our framework with empirical results. 
We first present ablation studies for better understanding of the source of gains with different components of the framework. Next, we quantify the refinement accuracy and show the efficacy of the proposed refinement block, as one of the key components of the proposed framework.
Finally, we present sensitivity analyses on two key hyperparameters of {\algname}, as it is crucial for an unsupervised AD method not to be sensitive to its hyperparameters.

\subsection{Ablation studies}
To shed more light on the source of gain coming from different components, we present ablation studies. Specifically, we include comparisons with the final ensemble model on the converged self-supervised extractor ({\algname} without the final OCC), without using data refinement but only employing an ensemble of OCCs, applying data refinement only for OCCs (but not for the representation updates), and data refinement with majority voting ({\algname} with majority voting).

\begin{table}[h!]
    \caption{Ablation studies. SOTA OCC methods are CutPaste \citep{li2021cutpaste} for MVTec dataset and \citep{sohn2020learning} for CIFAR-10 dataset with 6\% noise. Metrics are (AUC/AP).}
    \centering
    \resizebox{0.99\textwidth}{!}{
    \begin{tabular}{c|c|c}
        \toprule
        Methods / Datasets & MVTec & CIFAR-10 \\
        \midrule
        SOTA OCC & 0.905$\pm$0.024 / 0.845$\pm$0.035 & 0.855$\pm$0.017 / 0.585$\pm$0.063 \\
        {\algname} without final OCC & 0.922$\pm$0.022 / 0.870$\pm$0.032 & 0.890$\pm$0.007 / 0.677$\pm$0.019 \\
        {\algname} without data refinement (Ensemble of OCCs) & 0.911$\pm$0.023 / 0.849$\pm$0.037 & 0.862$\pm$0.010 / 0.599$\pm$0.031 \\ 
        Data refinement for only OCC & 0.918$\pm$0.028 / 0.858$\pm$0.039 & 0.885$\pm$0.006 / 0.644$\pm$0.016 \\
        {\algname} with majority voting & 0.925$\pm$0.017 / 0.873$\pm$0.026  & 0.893$\pm$0.005 / 0.675$\pm$0.014 \\
        \midrule
        {\algname} & \textbf{0.937$\pm$0.018 / 0.887$\pm$0.032} & \textbf{0.910$\pm$0.005 / 0.709$\pm$0.013} \\
        \bottomrule
    \end{tabular}
    }
    \label{tab:additional_ablation_study}
\end{table}

In Table.~\ref{tab:additional_ablation_study}, the proposed version of {\algname} (with an additional final OCC) outperforms the version without the final OCC. Because while fitting the individual OCC models in the ensemble, we do not exclude the possible anomaly samples for diversity of the trained submodels, so the anomaly decision boundaries can be fitted robustly. Ensemble of one-class classifiers is employed to identify the possible anomalous samples in the training set rather than yielding final anomaly score predictions. Therefore, we do not exclude the possible anomalous samples to train the weak one-class classifier. In that regard, if we directly utilized the outputs of the ensemble for the final anomaly scores, the performance would be worse (as shown in Table~\ref{tab:additional_ablation_study}).
Also, employment of ensemble of OCCs without data refinement, yields worse performance than the proposed ({\algname}), underlining the importance of the core data refinement idea of {\algname}, which was further eluded in Sec. \ref{sec:exp_ablation_refinement_efficacy}. Lastly, as opposed to applying data refinement only for OCCs, we find that learning representation with refined data plays a crucial role, resulting in another improvement compared to {\algname} using fixed representations trained on the entire dataset. The proposed version of {\algname} achieved statistically significant improvements over alternatives with CIFAR-10 datasets in terms of both AUC and AP but not with MVTec datasets due to their small sample size per category (100s-500s).



\subsection{Refinement efficacy}
\label{sec:exp_ablation_refinement_efficacy}

\begin{figure*}[h!]
\centering
\begin{subfigure}[b]{0.24\textwidth}
  \includegraphics[width=\linewidth]{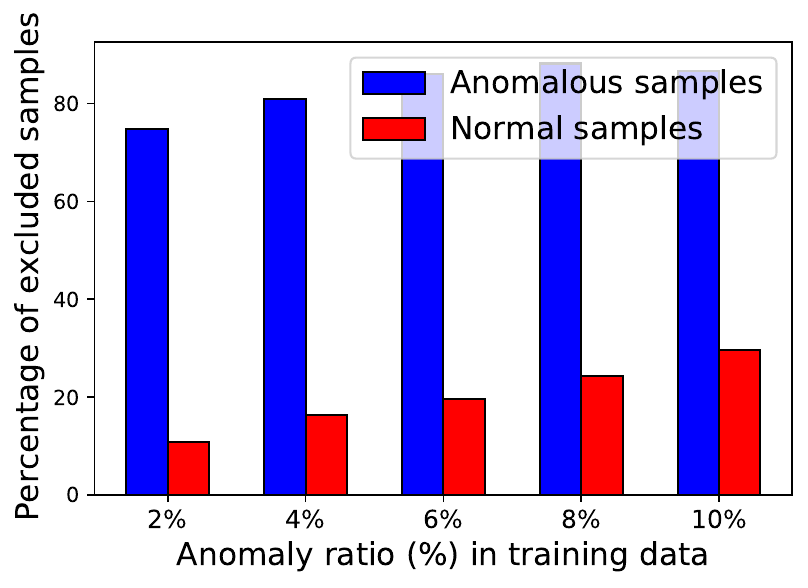}
  \caption{CIFAR-10}
\end{subfigure}
\begin{subfigure}[b]{0.24\textwidth}
  \includegraphics[width=\linewidth]{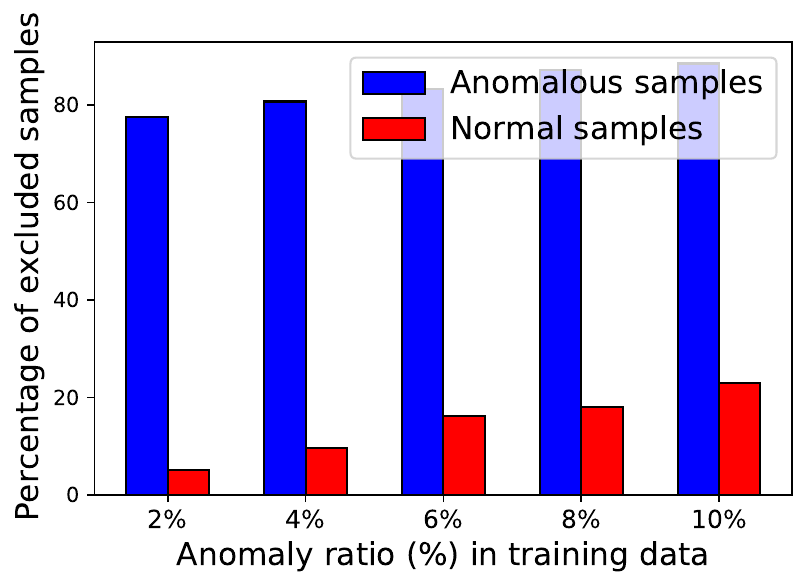}
  \caption{MVTec}
\end{subfigure}
\begin{subfigure}[b]{0.24\textwidth}
  \includegraphics[width=\linewidth]{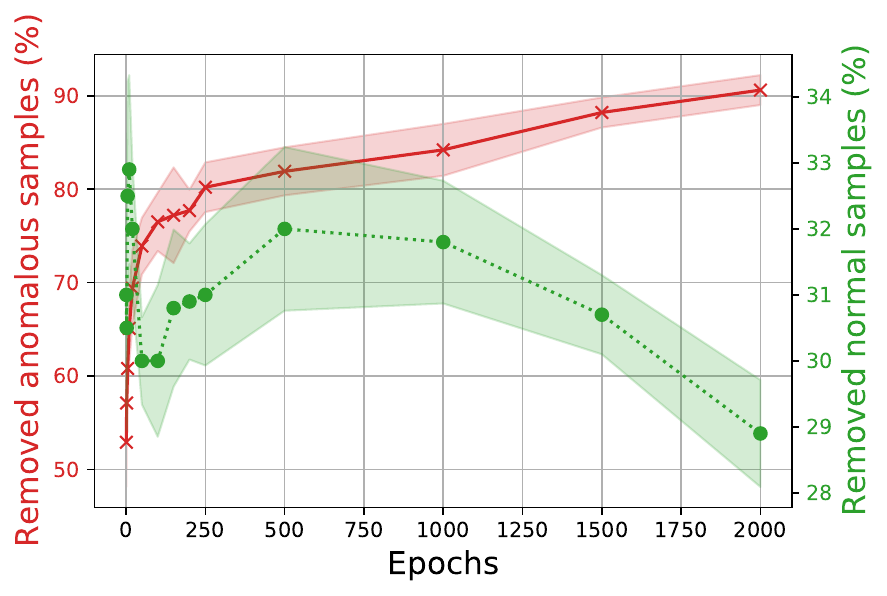}
  \caption{CIFAR-10}
\end{subfigure}
\begin{subfigure}[b]{0.24\textwidth}
  \includegraphics[width=\linewidth]{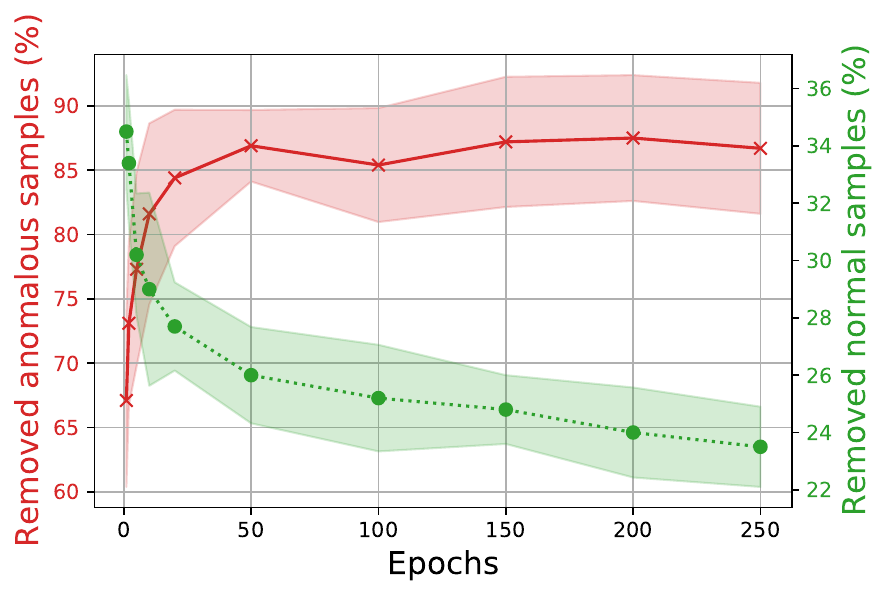}
  \caption{MVTec}
\end{subfigure}
\caption{Percentage of excluded anomalous and normal samples by data refinement (a, b) with different anomaly ratios in training data and (c, d) over training epochs with 10\% anomaly ratio.}\label{fig:selection_number_result}
\end{figure*}

It is crucial for the proposed data refinement block being accurate in filtering the correct anomaly samples, to minimize the impact of contamination. 
We evaluate the accuracy of this filtering of the proposed data refinement block. 
As Fig.~\ref{fig:selection_number_result}(a, b) shows, with data refinement, we can exclude more than 80\% of anomalies in the training set without removing too many normal samples. 
For example, among 4\% anomalies in CIFAR-10 data, {\algname} is able to exclude 80\% anomalies while removing less than 20\% normal samples.
Such a high recall of anomalies of {\algname} is not only useful for unsupervised AD, but can also help improving the annotation efficiency when a budget for active learning is available. 
Fig.~\ref{fig:selection_number_result}(c, d) demonstrates the removed normal and abnormal samples by the data refinement module over training epochs. 
It shows that better representation learning (as the number of training epochs increases) consistently improves the efficacy of the data refinement. This positive reinforcement between better representation learning and higher filtering accuracy, constitute one key aspect of the proposed framework.

\subsection{Sensitivity to hyperparameters}
\label{sec:exp_ablation_sensitivity}

\begin{figure*}[h!]
\centering
\begin{subfigure}[b]{0.33\textwidth}
  \includegraphics[width=\linewidth]{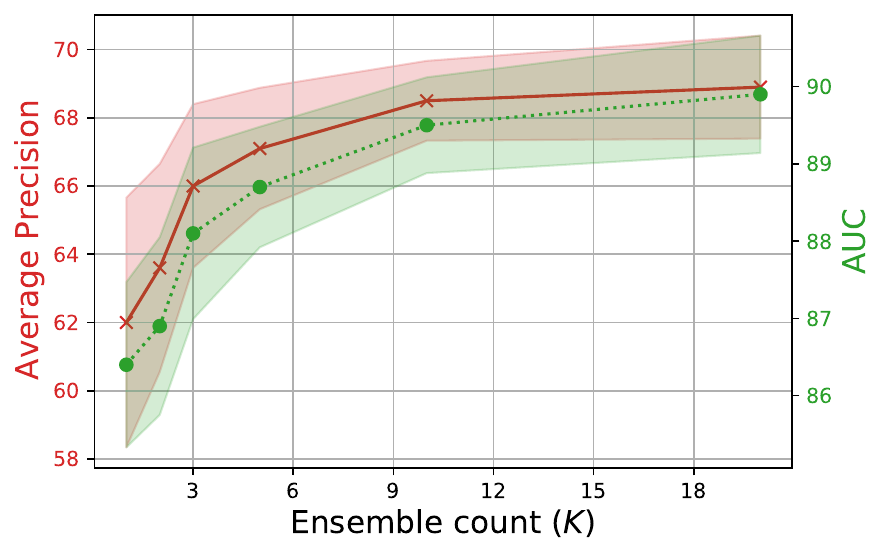}
  \includegraphics[width=\linewidth]{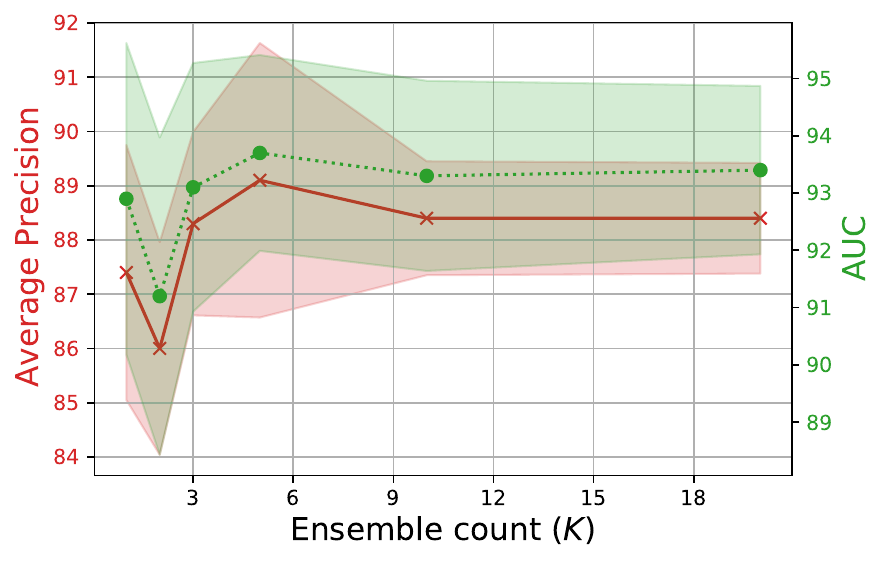}
  \caption{Ensemble count $K$}
  \label{fig:ablation_result_k}
\end{subfigure}
\begin{subfigure}[b]{0.33\textwidth}
  \includegraphics[width=\linewidth]{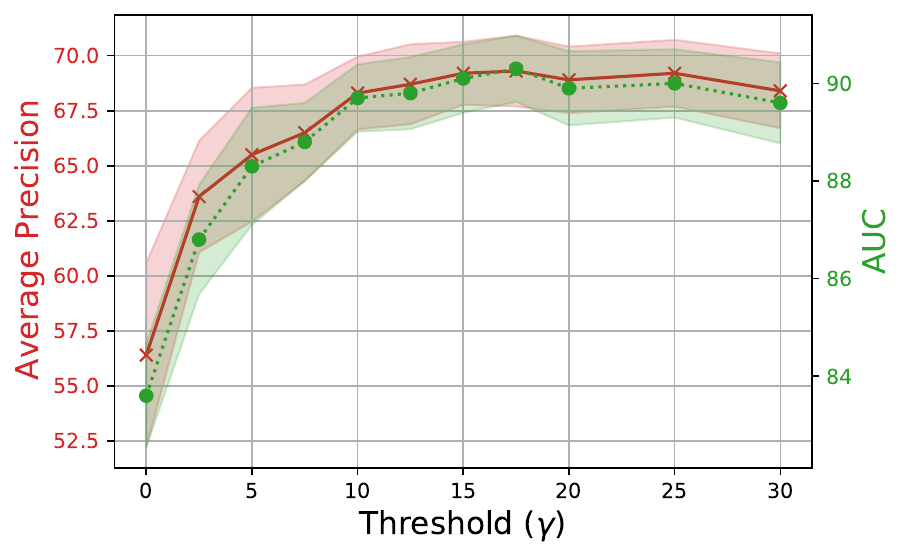}
  \includegraphics[width=\linewidth]{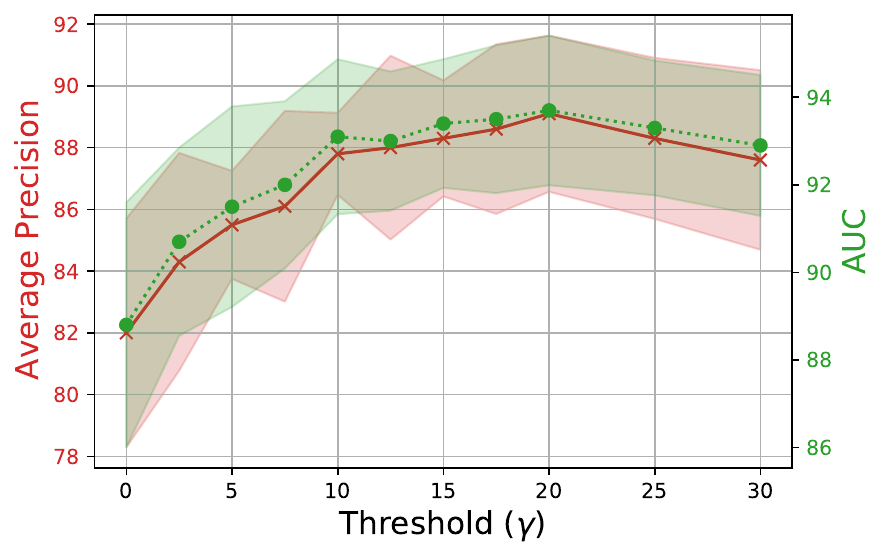}
  \caption{Percentile threshold $\gamma$}
  \label{fig:ablation_result_gamma}
\end{subfigure}
\caption{On (top) CIFAR-10 and (bottom) MVTec under 10\% anomaly ratio setting, sensitivity analyses for the (a) ensemble count $K$, (b) percentile threshold $\gamma$.}\label{fig:ablation_result}
\end{figure*}

{\algname} is designed for fully-unsupervised AD, thus, it is important to ensure robust performance against changes in the hyperparameters, as model selection without labeled data would be very challenging.
To demonstrate this, we conduct sensitivity analyses for the key hyperparameters, shown in Fig.~\ref{fig:ablation_result}. 

Fig.~\ref{fig:ablation_result_k} shows the impact of the number of OCCs in the ensemble, $K$. In general, the performance saturates even after a small number of OCCs, and a typical value of $K$ around $\sim$10 seems applicable across most datasets. On CIFAR-10 (top in Fig.~\ref{fig:ablation_result_k}), we observe slight improvements with higher $K$ as the number of samples to train each OCC ($N/K$) would be sufficient. 

Fig.~\ref{fig:ablation_result_gamma}, shows the impact of the threshold parameter, $\gamma$. Overall, an intermediate value seems optimal and the performance does not change much around it. We observe that {\algname} performs robustly when $\gamma$ is set to be larger than the actual anomaly ratio (10\%). 
When $\gamma$ is less than 10\%, however, we observe some drop in performance as $\gamma$ decreases. 
Yet, we note that {\algname} still improves over alternatives regardless of the threshold. 
The analyses suggest that $\gamma$ could be set to be anywhere from the true anomaly ratio and and 2$\times$ the anomaly ratio to maximize its effectiveness. 
For the scenario of unknown true anomaly ratio, we propose Otsu's method (see Sec.~\ref{sect:otsu_method}).

\section{{\algname} with unknown anomaly ratio: Otsu's method}\label{sect:otsu_method}

Proposed AD framework ({\algname}) is significantly better than alternatives in fully-unsupervised setting, as shown in wide range of scenarios. Despite being in fully-unsupervised setting, one information reliance of {\algname} is the true anomaly ratio. In practice, many applications have a good estimate for it from domain knowledge -- e.g. banks know typical credit card fraud ratio. Yet, there could be some applications that the true anomaly ratio may also be completely unknown. In this section, we propose an extension of {\algname} to address this scenario -- performing unsupervised AD without any information on anomaly ratio.

A key component would be automatic threshold selection. Some previous works \citep{xia2015learning,beggel2019robust,pang2020self} studied metrics like intra-class variance minimization for determining the threshold for data refinement. 
In a similar vein, we propose integrating Otsu’s method \citep{sezgin2004survey} into {\algname} to identify the threshold between normal and anomalous samples. We use this threshold for selecting the hyper-parameter ($\gamma$) instead of twice of the true anomaly ratio. The key idea of the Otsu’s method is aiming to find the threshold that minimizes the intra-class variance. This is defined as the weighted sum of variances of the two classes \footnote{More details can be found here \url{https://en.wikipedia.org/wiki/Otsu\%27s_method}}. Let us denote the normality scores as $\{s_i\}_{i=1}^N$ and threshold as $\eta$. Then, we pick the threshold ($\eta$) that minimizes the weighted sum of the variance $(w_0(\eta) \times \sigma_0(\eta)+w_1(\eta) \times \sigma_1(\eta))$ where
$w_0(\eta) = \sum_{i=1}^N \mathbb{I}(s_i < \eta) / N$ and $w_1(\eta) = \sum_{i=1}^N \mathbb{I}(s_i \geq \eta) / N$. $\sigma_0(\eta)$ and $\sigma_1(\eta)$ are the variances of each class. The optimal threshold ($\eta^*$) is determined as 
\begin{equation}
\eta^* = \min_\eta w_0(\eta) \times \sigma_0(\eta)+w_1(\eta) \times \sigma_1(\eta).
\end{equation} 
We use the twice of $\eta^*$ as the hyperparameter ($\gamma$) in {\algname} based on the sensitivity analyses described in Sec.~\ref{sec:exp_ablation_sensitivity}.

\begin{figure}
\centering
\includegraphics[width=0.4\linewidth]{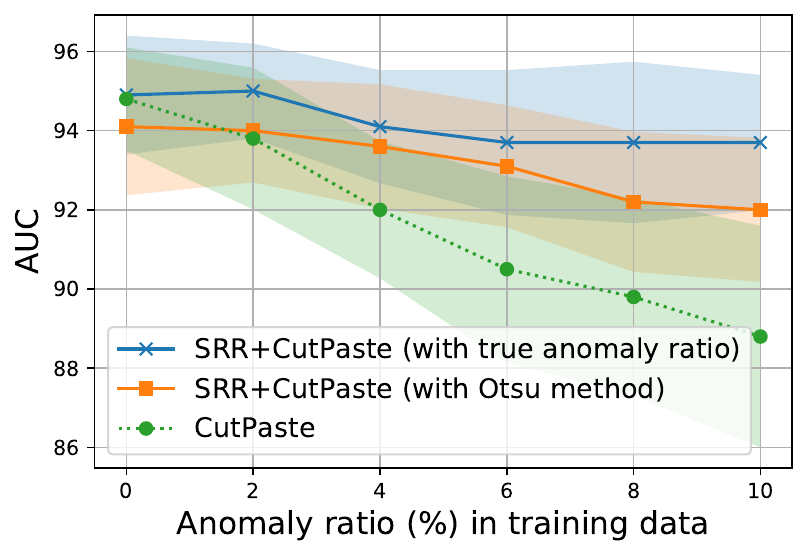}
\caption{Unsupervised AD performance for {\algname} with Otsu's method on MVTec dataset in AUC.}\label{fig:otsu_result}
\end{figure}

\begin{table}[h!]
    \centering
    \caption{Performances with Otsu's method. SOTA OCC methods are \citep{sohn2020learning} for CIFAR-10 dataset and GOAD \citep{bergman2020classification} for Thyroid dataset. We introduce 6\% noise on MVTec dataset and 1.5\% noise on the Thyroid dataset. Metrics are (AUC/AP) for CIFAR-10 dataset and F1 score for Thyroid dataset.}
    \begin{tabular}{c|c|c}
        \toprule
        Methods / Datasets & CIFAR-10 & Thyroid \\
        \midrule
        SOTA OCC & 0.855 / 0.585 & 0.506 \\
        {\algname} with Otsu's method & 0.906 / 0.703 & 0.623 \\
        {\algname} & \textbf{0.910 / 0.709} & \textbf{0.639} \\
        \bottomrule
    \end{tabular}
    \label{tab:additional_otsu}
\end{table}

We evaluate the performance of {\algname} with Otsu's method and compare to the state-of-the-art OCC \citep{li2021cutpaste} and the original {\algname} with the knowledge of the true anomaly ratio on MVTec dataset.
Fig.~\ref{fig:otsu_result} demonstrates that even without true anomaly ratio, the performance of {\algname} can be significantly better than the state-of-the-art OCC \citep{li2021cutpaste} with Otsu's method. Per results, the knowledge of true anomaly ratio is definitely a useful information for maximizing the performance of {\algname} in fully unsupervised settings.
We further extend the experimental results using Otsu's method with other datasets such as CIFAR-10 and Thyroid. Similarly, Table~\ref{tab:additional_otsu} show that Otsu’s method yields only slight degradation compared to {\algname} with true anomaly ratio; however, it still significantly outperforms SOTA OCC baselines.

\section{Conclusion}\label{sec:conclusion}
AD has wide range of use cases with significant importance in real world applications, from detecting security threats to the financial system to identifying faulty behaviors of manufacturing machines. 
A challenging and costly aspect of building an AD system is that anomalies are rare and not easily detectable by humans, yielding high complexity and cost. 
To this end, we propose, {\algname}, a canonical AD framework to enable high performance AD without any labels. 
{\algname} employs an ensemble of multiple OCCs to propose candidate anomaly samples that are refined from training, which allows more robust fitting of the anomaly decision boundaries as well as better learning of data representations. 
{\algname} can be flexibly integrated with any OCC, and applied on raw data or on trainable representations.
We demonstrate the state-of-the-art AD performance of {\algname} on multiple tabular and image datasets from various applications. We provide detailed analyses on the key contributing factors of {\algname}, which we hope to provide further guidance in AD research. Lastly, we extend {\algname} to the scenario of not possessing any information on the anomaly ratio. We leave some important aspects, explainability and reliability of AD, to future work. 

\bibliography{tmlr_srr}

\begin{thebibliography}{58}
\providecommand{\natexlab}[1]{#1}
\providecommand{\url}[1]{\texttt{#1}}
\expandafter\ifx\csname urlstyle\endcsname\relax
  \providecommand{\doi}[1]{doi: #1}\else
  \providecommand{\doi}{doi: \begingroup \urlstyle{rm}\Url}\fi

\bibitem[Akcay et~al.(2018)Akcay, Atapour-Abarghouei, and
  Breckon]{akcay2018ganomaly}
Samet Akcay, Amir Atapour-Abarghouei, and Toby~P Breckon.
\newblock Ganomaly: Semi-supervised anomaly detection via adversarial training.
\newblock In \emph{ACCV}, 2018.

\bibitem[Asuncion \& Newman(2007)Asuncion and Newman]{asuncion2007uci}
Arthur Asuncion and David Newman.
\newblock {UCI} machine learning repository, 2007.

\bibitem[Barua et~al.(2012)Barua, Islam, Yao, and Murase]{barua2012mwmote}
Sukarna Barua, Md~Monirul Islam, Xin Yao, and Kazuyuki Murase.
\newblock Mwmote--majority weighted minority oversampling technique for
  imbalanced data set learning.
\newblock \emph{IEEE Trans on knowledge and data engineering}, 26\penalty0
  (2):\penalty0 405--425, 2012.

\bibitem[Beggel et~al.(2019)Beggel, Pfeiffer, and Bischl]{beggel2019robust}
Laura Beggel, Michael Pfeiffer, and Bernd Bischl.
\newblock Robust anomaly detection in images using adversarial autoencoders.
\newblock \emph{Machine Learning and Knowledge Discovery in Databases}, 2019.

\bibitem[Berg et~al.(2019)Berg, Ahlberg, and Felsberg]{berg2019unsupervised}
Amanda Berg, J{\"o}rgen Ahlberg, and Michael Felsberg.
\newblock Unsupervised learning of anomaly detection from contaminated image
  data using simultaneous encoder training.
\newblock \emph{arXiv preprint arXiv:1905.11034}, 2019.

\bibitem[Bergman \& Hoshen(2019)Bergman and Hoshen]{bergman2020classification}
Liron Bergman and Yedid Hoshen.
\newblock Classification-based anomaly detection for general data.
\newblock In \emph{International Conference on Learning Representations}, 2019.

\bibitem[Bergmann et~al.(2019)Bergmann, Fauser, Sattlegger, and
  Steger]{bergmann2019mvtec}
Paul Bergmann, Michael Fauser, David Sattlegger, and Carsten Steger.
\newblock {MVTec AD}--a comprehensive real-world dataset for unsupervised
  anomaly detection.
\newblock In \emph{CVPR}, 2019.

\bibitem[Berthelot et~al.(2019)Berthelot, Carlini, Goodfellow, Papernot,
  Oliver, and Raffel]{berthelot2019mixmatch}
David Berthelot, Nicholas Carlini, Ian Goodfellow, Nicolas Papernot, Avital
  Oliver, and Colin~A Raffel.
\newblock Mixmatch: A holistic approach to semi-supervised learning.
\newblock \emph{Advances in Neural Information Processing Systems}, 32, 2019.

\bibitem[Blanchard et~al.(2010)Blanchard, Lee, and Scott]{blanchard2010semi}
Gilles Blanchard, Gyemin Lee, and Clayton Scott.
\newblock Semi-supervised novelty detection.
\newblock \emph{JMLR}, 11:\penalty0 2973--3009, 2010.

\bibitem[Blum \& Mitchell(1998)Blum and Mitchell]{blum1998combining}
Avrim Blum and Tom Mitchell.
\newblock Combining labeled and unlabeled data with co-training.
\newblock In \emph{CLT}, 1998.

\bibitem[Branco et~al.(2015)Branco, Torgo, and Ribeiro]{branco2015survey}
Paula Branco, Luis Torgo, and Rita Ribeiro.
\newblock A survey of predictive modelling under imbalanced distributions.
\newblock \emph{arXiv preprint arXiv:1505.01658}, 2015.

\bibitem[Breunig et~al.(2000)Breunig, Kriegel, Ng, and Sander]{breunig2000lof}
Markus~M Breunig, Hans-Peter Kriegel, Raymond~T Ng, and J{\"o}rg Sander.
\newblock Lof: identifying density-based local outliers.
\newblock In \emph{International conference on management of data}, 2000.

\bibitem[Brodley et~al.(1996)Brodley, Friedl, et~al.]{brodley1996identifying}
Carla~E Brodley, Mark~A Friedl, et~al.
\newblock Identifying and eliminating mislabeled training instances.
\newblock In \emph{Proc National Conference on Artificial Intelligence}, 1996.

\bibitem[Cand{\`e}s et~al.(2011)Cand{\`e}s, Li, Ma, and
  Wright]{candes2011robust}
Emmanuel~J Cand{\`e}s, Xiaodong Li, Yi~Ma, and John Wright.
\newblock Robust principal component analysis?
\newblock \emph{Journal of the ACM (JACM)}, 58\penalty0 (3):\penalty0 1--37,
  2011.

\bibitem[Chawla et~al.(2002)Chawla, Bowyer, Hall, and
  Kegelmeyer]{chawla2002smote}
Nitesh~V Chawla, Kevin~W Bowyer, Lawrence~O Hall, and W~Philip Kegelmeyer.
\newblock Smote: synthetic minority over-sampling technique.
\newblock \emph{Journal of artificial intelligence research}, 16:\penalty0
  321--357, 2002.

\bibitem[Chen et~al.(2020)Chen, Kornblith, Norouzi, and Hinton]{chen2020simple}
Ting Chen, Simon Kornblith, Mohammad Norouzi, and Geoffrey Hinton.
\newblock A simple framework for contrastive learning of visual
  representations.
\newblock In \emph{ICML}, 2020.

\bibitem[Elson et~al.(2007)Elson, Douceur, Howell, and Saul]{elson2007asirra}
Jeremy Elson, John~R Douceur, Jon Howell, and Jared Saul.
\newblock Asirra: a captcha that exploits interest-aligned manual image
  categorization.
\newblock In \emph{Proc Computer and Communications Security}, 2007.

\bibitem[Estabrooks et~al.(2004)Estabrooks, Jo, and
  Japkowicz]{estabrooks2004multiple}
Andrew Estabrooks, Taeho Jo, and Nathalie Japkowicz.
\newblock A multiple resampling method for learning from imbalanced data sets.
\newblock \emph{Computational intelligence}, 20\penalty0 (1):\penalty0 18--36,
  2004.

\bibitem[Feng et~al.(2021)Feng, Hong, and Zheng]{feng2021mist}
Jia-Chang Feng, Fa-Ting Hong, and Wei-Shi Zheng.
\newblock Mist: Multiple instance self-training framework for video anomaly
  detection.
\newblock In \emph{CVPR}, 2021.

\bibitem[Galar et~al.(2011)Galar, Fernandez, Barrenechea, Bustince, and
  Herrera]{galar2011review}
Mikel Galar, Alberto Fernandez, Edurne Barrenechea, Humberto Bustince, and
  Francisco Herrera.
\newblock A review on ensembles for the class imbalance problem: bagging-,
  boosting-, and hybrid-based approaches.
\newblock \emph{IEEE Trans on Systems, Man, and Cybernetics}, 42\penalty0
  (4):\penalty0 463--484, 2011.

\bibitem[Golan \& El-Yaniv(2018)Golan and El-Yaniv]{golan2018deep}
Izhak Golan and Ran El-Yaniv.
\newblock Deep anomaly detection using geometric transformations.
\newblock In \emph{Proceedings of the 32nd International Conference on Neural
  Information Processing Systems}, pp.\  9781--9791, 2018.

\bibitem[G{\"o}rnitz et~al.(2013)G{\"o}rnitz, Kloft, Rieck, and
  Brefeld]{gornitz2013toward}
Nico G{\"o}rnitz, Marius Kloft, Konrad Rieck, and Ulf Brefeld.
\newblock Toward supervised anomaly detection.
\newblock \emph{Journal of Artificial Intelligence Research}, 46:\penalty0
  235--262, 2013.

\bibitem[Han et~al.(2018)Han, Yao, Yu, Niu, Xu, Hu, Tsang, and
  Sugiyama]{han2018co}
Bo~Han, Quanming Yao, Xingrui Yu, Gang Niu, Miao Xu, Weihua Hu, Ivor~W Tsang,
  and Masashi Sugiyama.
\newblock Co-teaching: robust training of deep neural networks with extremely
  noisy labels.
\newblock In \emph{NIPS}, 2018.

\bibitem[Hendrycks et~al.(2018)Hendrycks, Mazeika, and
  Dietterich]{hendrycks2018deep}
Dan Hendrycks, Mantas Mazeika, and Thomas Dietterich.
\newblock Deep anomaly detection with outlier exposure.
\newblock In \emph{International Conference on Learning Representations}, 2018.

\bibitem[Hwang et~al.(2011)Hwang, Park, and Kim]{hwang2011new}
Jae~Pil Hwang, Seongkeun Park, and Euntai Kim.
\newblock A new weighted approach to imbalanced data classification problem via
  support vector machine with quadratic cost function.
\newblock \emph{Expert Systems with Applications}, 38\penalty0 (7):\penalty0
  8580--8585, 2011.

\bibitem[Komodakis \& Gidaris(2018)Komodakis and
  Gidaris]{gidaris2018unsupervised}
Nikos Komodakis and Spyros Gidaris.
\newblock Unsupervised representation learning by predicting image rotations.
\newblock In \emph{International Conference on Learning Representations
  (ICLR)}, 2018.

\bibitem[Krizhevsky \& Hinton(2009)Krizhevsky and
  Hinton]{krizhevsky2009learning}
Alex Krizhevsky and Geoffrey Hinton.
\newblock Learning multiple layers of features from tiny images.
\newblock 2009.

\bibitem[Lai et~al.(2019)Lai, Zou, and Lerman]{lai2019robust}
Chieh-Hsin Lai, Dongmian Zou, and Gilad Lerman.
\newblock Robust subspace recovery layer for unsupervised anomaly detection.
\newblock In \emph{International Conference on Learning Representations}, 2019.

\bibitem[Latecki et~al.(2007)Latecki, Lazarevic, and
  Pokrajac]{latecki2007outlier}
Longin~Jan Latecki, Aleksandar Lazarevic, and Dragoljub Pokrajac.
\newblock Outlier detection with kernel density functions.
\newblock In \emph{International Workshop on Machine Learning and Data Mining
  in Pattern Recognition}, 2007.

\bibitem[Lee(2000)]{lee2000noisy}
Sauchi~Stephen Lee.
\newblock Noisy replication in skewed binary classification.
\newblock \emph{Computational statistics \& data analysis}, 34\penalty0
  (2):\penalty0 165--191, 2000.

\bibitem[Li et~al.(2021)Li, Sohn, Yoon, and Pfister]{li2021cutpaste}
Chun-Liang Li, Kihyuk Sohn, Jinsung Yoon, and Tomas Pfister.
\newblock Cutpaste: Self-supervised learning for anomaly detection and
  localization.
\newblock In \emph{CVPR}, 2021.

\bibitem[Li et~al.(2019)Li, Socher, and Hoi]{li2019dividemix}
Junnan Li, Richard Socher, and Steven~CH Hoi.
\newblock Dividemix: Learning with noisy labels as semi-supervised learning.
\newblock In \emph{ICLR}, 2019.

\bibitem[Liu et~al.(2007)Liu, Ghosh, and Martin]{liu2007generative}
Alexander Liu, Joydeep Ghosh, and Cheryl~E Martin.
\newblock Generative oversampling for mining imbalanced datasets.
\newblock In \emph{DMIN}, pp.\  66--72, 2007.

\bibitem[Liu et~al.(2008)Liu, Ting, and Zhou]{liu2008isolation}
Fei~Tony Liu, Kai~Ming Ting, and Zhi-Hua Zhou.
\newblock Isolation forest.
\newblock In \emph{ICDM}, 2008.

\bibitem[McLachlan(1975)]{mclachlan1975iterative}
Geoffrey~J McLachlan.
\newblock Iterative reclassification procedure for constructing an
  asymptotically optimal rule of allocation in discriminant analysis.
\newblock \emph{Journal of the American Statistical Association}, 70\penalty0
  (350):\penalty0 365--369, 1975.

\bibitem[Meng et~al.(2021)Meng, Wang, Liang, Yao, Zhou, and
  Zhang]{meng2021semi}
Xuying Meng, Suhang Wang, Zhimin Liang, Di~Yao, Jihua Zhou, and Yujun Zhang.
\newblock Semi-supervised anomaly detection in dynamic communication networks.
\newblock \emph{Information Sciences}, 571:\penalty0 527--542, 2021.

\bibitem[Min et~al.(2018)Min, Long, Liu, Cui, Cai, and Ma]{min2018ids}
Erxue Min, Jun Long, Qiang Liu, Jianjing Cui, Zhiping Cai, and Junbo Ma.
\newblock Su-ids: A semi-supervised and unsupervised framework for network
  intrusion detection.
\newblock In \emph{International Conference on Cloud Computing and Security},
  2018.

\bibitem[Mohseni et~al.(2021)Mohseni, Yap, Yolland, Koochek, and
  Atkins]{mohseni2021can}
Mohammadreza Mohseni, Jordan Yap, William Yolland, Arash Koochek, and Stella
  Atkins.
\newblock Can self-training identify suspicious ugly duckling lesions?
\newblock In \emph{CVPR}, 2021.

\bibitem[M{\~u}noz-Mar{\'\i} et~al.(2010)M{\~u}noz-Mar{\'\i}, Bovolo,
  G{\'o}mez-Chova, Bruzzone, and Camp-Valls]{munoz2010semisupervised}
Jordi M{\~u}noz-Mar{\'\i}, Francesca Bovolo, Luis G{\'o}mez-Chova, Lorenzo
  Bruzzone, and Gustavo Camp-Valls.
\newblock Semisupervised one-class support vector machines for classification
  of remote sensing data.
\newblock \emph{IEEE transactions on geoscience and remote sensing},
  48\penalty0 (8):\penalty0 3188--3197, 2010.

\bibitem[Pang et~al.(2020)Pang, Yan, Shen, Hengel, and Bai]{pang2020self}
Guansong Pang, Cheng Yan, Chunhua Shen, Anton van~den Hengel, and Xiao Bai.
\newblock Self-trained deep ordinal regression for end-to-end video anomaly
  detection.
\newblock In \emph{CVPR}, 2020.

\bibitem[Ren et~al.(2019)Ren, Liu, Fertig, Snoek, Poplin, Depristo, Dillon, and
  Lakshminarayanan]{ren2019likelihood}
Jie Ren, Peter~J Liu, Emily Fertig, Jasper Snoek, Ryan Poplin, Mark Depristo,
  Joshua Dillon, and Balaji Lakshminarayanan.
\newblock Likelihood ratios for out-of-distribution detection.
\newblock \emph{Advances in neural information processing systems}, 32, 2019.

\bibitem[Reynolds(2009)]{reynolds2009gaussian}
Douglas~A Reynolds.
\newblock Gaussian mixture models.
\newblock \emph{Encyclopedia of biometrics}, 741:\penalty0 659--663, 2009.

\bibitem[Ruff et~al.(2018)Ruff, Vandermeulen, Goernitz, Deecke, Siddiqui,
  Binder, M{\"u}ller, and Kloft]{ruff2018deep}
Lukas Ruff, Robert Vandermeulen, Nico Goernitz, Lucas Deecke, Shoaib~Ahmed
  Siddiqui, Alexander Binder, Emmanuel M{\"u}ller, and Marius Kloft.
\newblock Deep one-class classification.
\newblock In \emph{ICML}, 2018.

\bibitem[Ruff et~al.(2020)Ruff, Vandermeulen, G{\"o}rnitz, Binder, M{\"u}ller,
  M{\"u}ller, and Kloft]{ruff2020deep}
Lukas Ruff, Robert~A Vandermeulen, Nico G{\"o}rnitz, Alexander Binder, Emmanuel
  M{\"u}ller, Klaus-Robert M{\"u}ller, and Marius Kloft.
\newblock Deep semi-supervised anomaly detection.
\newblock In \emph{ICLR}, 2020.

\bibitem[Sch{\"o}lkopf et~al.(1999)Sch{\"o}lkopf, Williamson, Smola,
  Shawe-Taylor, Platt, et~al.]{scholkopf1999support}
Bernhard Sch{\"o}lkopf, Robert~C Williamson, Alexander~J Smola, John
  Shawe-Taylor, John~C Platt, et~al.
\newblock Support vector method for novelty detection.
\newblock In \emph{NIPS}, 1999.

\bibitem[Scudder(1965)]{scudder1965probability}
Henry Scudder.
\newblock Probability of error of some adaptive pattern-recognition machines.
\newblock \emph{IEEE Transactions on Information Theory}, 11\penalty0
  (3):\penalty0 363--371, 1965.

\bibitem[Sezgin \& Sankur(2004)Sezgin and Sankur]{sezgin2004survey}
Mehmet Sezgin and B{\"u}lent Sankur.
\newblock Survey over image thresholding techniques and quantitative
  performance evaluation.
\newblock \emph{Journal of Electronic imaging}, 13\penalty0 (1):\penalty0
  146--165, 2004.

\bibitem[Sohn et~al.(2020)Sohn, Berthelot, Carlini, Zhang, Zhang, Raffel,
  Cubuk, Kurakin, and Li]{sohn2020fixmatch}
Kihyuk Sohn, David Berthelot, Nicholas Carlini, Zizhao Zhang, Han Zhang,
  Colin~A Raffel, Ekin~Dogus Cubuk, Alexey Kurakin, and Chun-Liang Li.
\newblock Fixmatch: Simplifying semi-supervised learning with consistency and
  confidence.
\newblock \emph{NeurIPS}, 2020.

\bibitem[Sohn et~al.(2021)Sohn, Li, Yoon, Jin, and Pfister]{sohn2020learning}
Kihyuk Sohn, Chun-Liang Li, Jinsung Yoon, Minho Jin, and Tomas Pfister.
\newblock Learning and evaluating representations for deep one-class
  classification.
\newblock In \emph{ICLR}, 2021.

\bibitem[Song et~al.(2017)Song, Jiang, Men, and Yang]{song2017hybrid}
Hongchao Song, Zhuqing Jiang, Aidong Men, and Bo~Yang.
\newblock A hybrid semi-supervised anomaly detection model for high-dimensional
  data.
\newblock \emph{Computational intelligence and neuroscience}, 2017.

\bibitem[Tax \& Duin(2004)Tax and Duin]{tax2004support}
David~MJ Tax and Robert~PW Duin.
\newblock Support vector data description.
\newblock \emph{Machine learning}, 54\penalty0 (1):\penalty0 45--66, 2004.

\bibitem[Wold et~al.(1987)Wold, Esbensen, and Geladi]{wold1987principal}
Svante Wold, Kim Esbensen, and Paul Geladi.
\newblock Principal component analysis.
\newblock \emph{Chemometrics and intelligent laboratory systems}, 2\penalty0
  (1-3):\penalty0 37--52, 1987.

\bibitem[Xia et~al.(2015)Xia, Cao, Wen, Hua, and Sun]{xia2015learning}
Yan Xia, Xudong Cao, Fang Wen, Gang Hua, and Jian Sun.
\newblock Learning discriminative reconstructions for unsupervised outlier
  removal.
\newblock In \emph{ICCV}, 2015.

\bibitem[Xiao et~al.(2017)Xiao, Rasul, and Vollgraf]{xiao2017fashion}
Han Xiao, Kashif Rasul, and Roland Vollgraf.
\newblock Fashion-mnist: a novel image dataset for benchmarking machine
  learning algorithms.
\newblock \emph{arXiv preprint arXiv:1708.07747}, 2017.

\bibitem[Xie et~al.(2020)Xie, Luong, Hovy, and Le]{xie2020self}
Qizhe Xie, Minh-Thang Luong, Eduard Hovy, and Quoc~V Le.
\newblock Self-training with noisy student improves imagenet classification.
\newblock In \emph{CVPR}, 2020.

\bibitem[Zhang \& Zuo(2008)Zhang and Zuo]{zhang2008learning}
Bangzuo Zhang and Wanli Zuo.
\newblock Learning from positive and unlabeled examples: A survey.
\newblock In \emph{2008 International Symposiums on Information Processing},
  2008.

\bibitem[Zhou \& Paffenroth(2017)Zhou and Paffenroth]{zhou2017anomaly}
Chong Zhou and Randy~C Paffenroth.
\newblock Anomaly detection with robust deep autoencoders.
\newblock In \emph{Proceedings of the 23rd ACM SIGKDD international conference
  on knowledge discovery and data mining}, pp.\  665--674, 2017.

\bibitem[Zong et~al.(2018)Zong, Song, Min, Cheng, Lumezanu, Cho, and
  Chen]{zong2018deep}
Bo~Zong, Qi~Song, Martin~Renqiang Min, Wei Cheng, Cristian Lumezanu, Daeki Cho,
  and Haifeng Chen.
\newblock Deep autoencoding gaussian mixture model for unsupervised anomaly
  detection.
\newblock In \emph{ICLR}, 2018.

\end{thebibliography}
\bibliographystyle{tmlr}

\newpage
\appendix
\section{Additional results}



\subsection{{\algname} on raw tabular features / learned image representations}\label{sect:raw_features}

\begin{figure*}[h!]
\centering
\begin{subfigure}[b]{0.27\textwidth}
  \includegraphics[width=\linewidth]{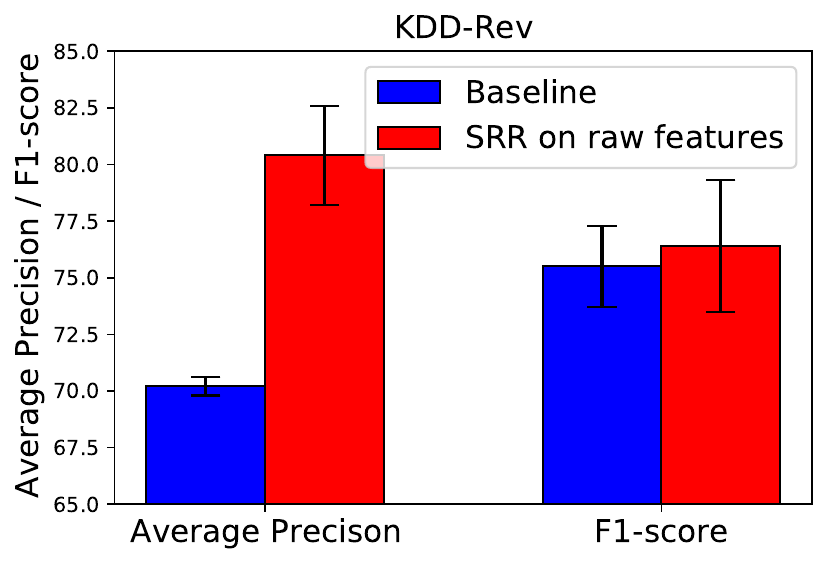}
  \caption{KDD-Rev}
\end{subfigure}
\begin{subfigure}[b]{0.27\textwidth}
  \includegraphics[width=\linewidth]{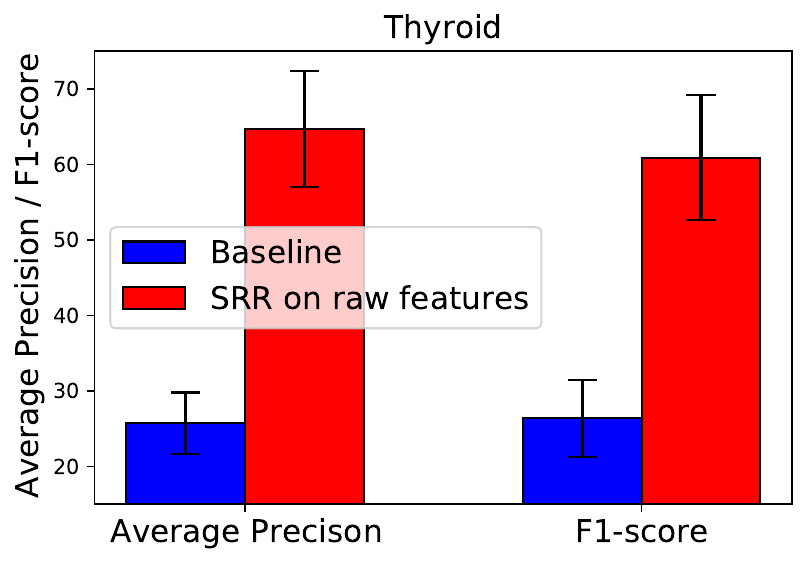}
  \caption{Thyroid}
\end{subfigure}
\begin{subfigure}[b]{0.27\textwidth}
  \includegraphics[width=\linewidth]{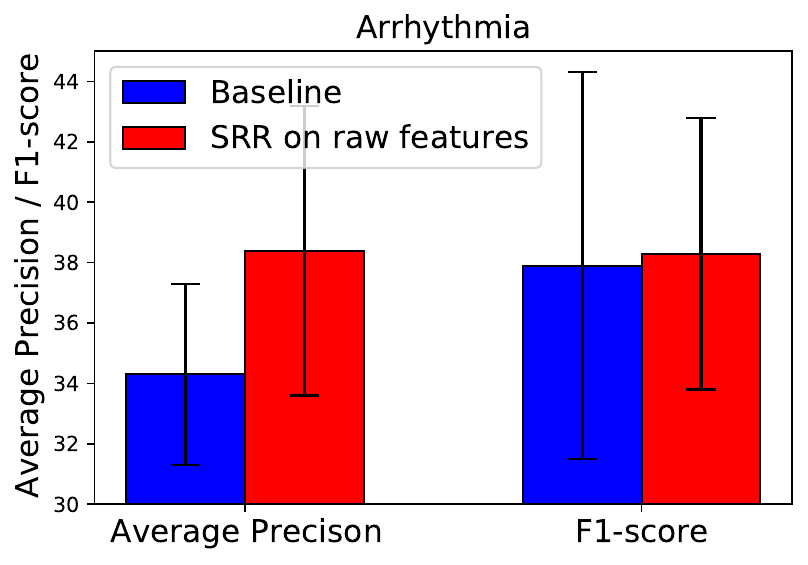}
  \caption{Arrhythmia}
\end{subfigure}\\
\begin{subfigure}[b]{0.27\textwidth}
  \includegraphics[width=\linewidth]{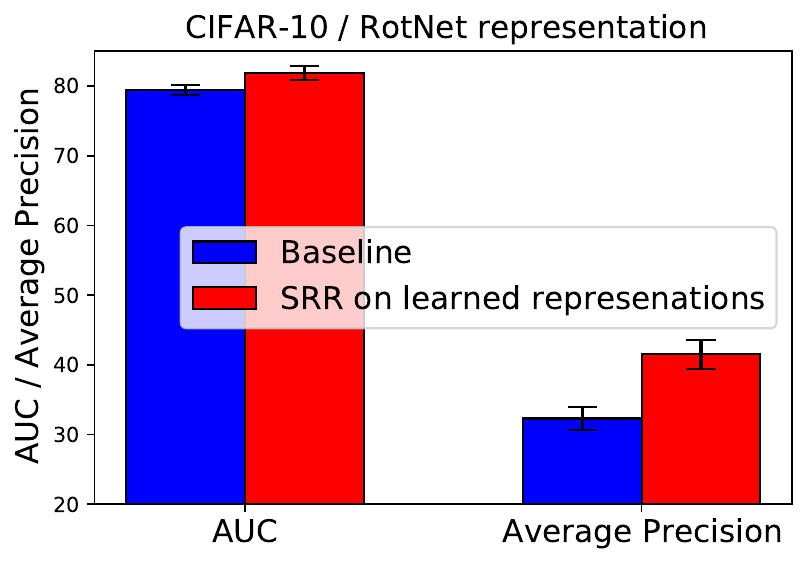}
  \caption{RotNet \citep{golan2018deep}}
\end{subfigure}
\begin{subfigure}[b]{0.27\textwidth}
  \includegraphics[width=\linewidth]{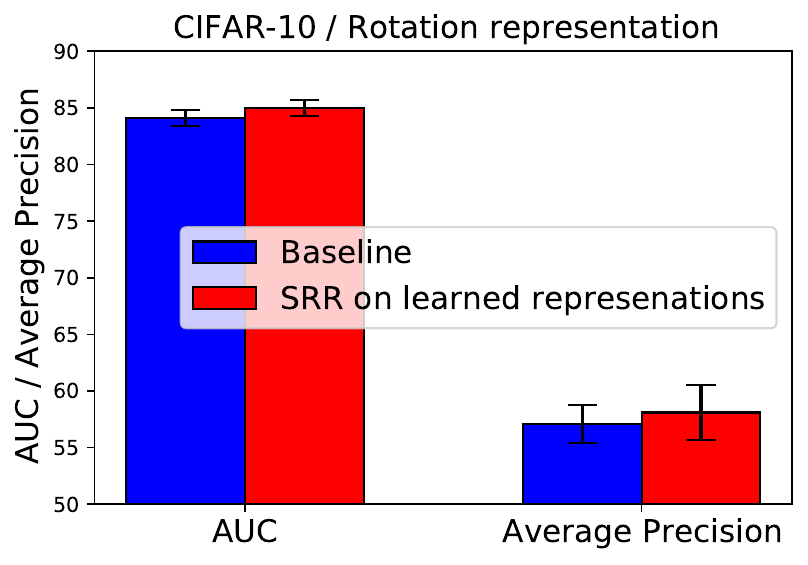}
  \caption{Rotation \citep{gidaris2018unsupervised}}
\end{subfigure}
\begin{subfigure}[b]{0.27\textwidth}
  \includegraphics[width=\linewidth]{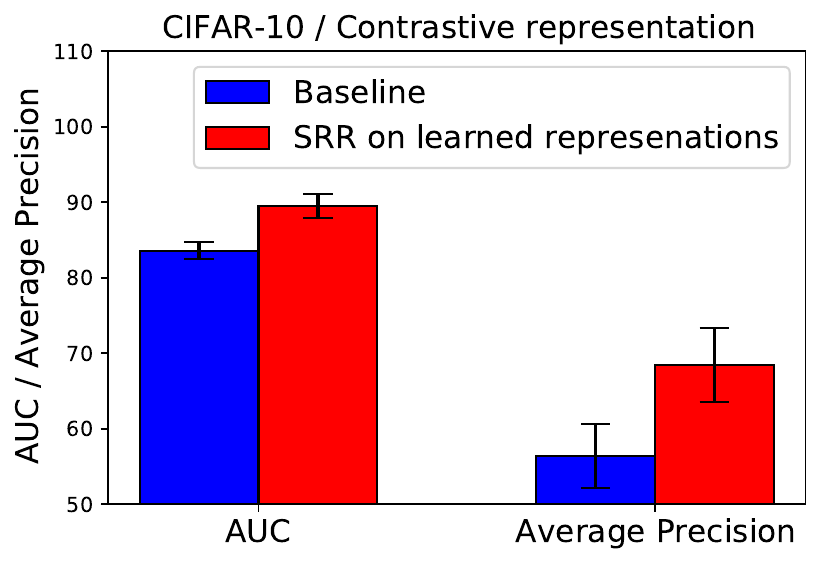}
  \caption{Contrastive \citep{sohn2020learning}}
\end{subfigure}
\caption{Performance of {\algname} on (top) raw tabular features and (lower) learned image representations. {\algname} consistently outperforms baseline and in some cases (e.g., Thyroid, and Contrastive \citep{sohn2020learning}), the performance improvements are significant. }\label{fig:raw}
\end{figure*}

{\algname} is also applicable on raw tabular features or learned image representation without representation update using data refinement. In this section, we demonstrate the performance improvements by {\algname} without representation update to verify the effectiveness of data refinement block of {\algname} for shallow OCCs.

Fig.~\ref{fig:raw} (upper) demonstrates consistent and significant performance improvements when we apply {\algname} on top of raw tabular features. Specifically, the Average Precision (AP) improvements are 10.2, 29.0, and 4.1 with KDD-Rev, Thyroid, and Arrhythmia tabular datasets, respectively.
We also apply {\algname} on top of various learned image representations. As can be seen in Fig.~\ref{fig:raw} (lower), the performance improvements of {\algname} are consistent across various different learned image representations (without representation update). For instance, the AP improvements are 9.2, 1.0, and 12.1 with learned image representations using RotNet \citep{golan2018deep}, Rotation \citep{gidaris2018unsupervised}, and Contrastive \citep{sohn2020learning}, respectively.

\subsection{Additional baselines}

\begin{table}[h!]
    \caption{Additional experiments with extra baselines from robust AD literature. We introduce 6\% noise on CIFAR-10 and KDD datasets. For Thyroid dataset, we introduce 1.5\% noise. Metrics are (AUC/AP) for image data and F1 score for tabular data.}
    \centering
    \begin{tabular}{c|c|c|c}
        \toprule
        Methods / Datasets & CIFAR-10 & Thyroid & KDD \\
        \midrule
        PCA & - & 0.299 & 0.836 \\
        Robust PCA & - & 0.377 & 0.893 \\
        LOF & - & 0.338 & 0.873 \\
        Robust AE & 0.636 / 0.174 & - & - \\
        {\algname} & \textbf{0.910 / 0.709} & \textbf{0.506} & \textbf{0.942} \\
        \bottomrule
    \end{tabular}
    \label{tab:additional_robust_baseline}
\end{table}

We add extra baselines from robust AD literature: Standard PCA \citep{wold1987principal}, Robust PCA \citep{candes2011robust} and Local Outlier Factor (LOF) \citep{breunig2000lof} for tabular data and Robust autoencoder (Robust AE) \citep{zhou2017anomaly} for image data. In Table~\ref{tab:additional_robust_baseline}, the performance of PCA and LOF are highly degraded even with a small amount of anomalies in the training data. For Robust PCA and Robust AE, the performance degradation is less but still significant in comparison to {\algname}. Overall, {\algname} outperforms other benchmarks in fully unsupervised settings, underlining the importance of data refinement in improving the robustness with contaminated data, as the core constituent of {\algname}.

\subsection{Convergence graphs}\label{sect:appendix_convergence}
\begin{figure}[h!]
\minipage{0.48\textwidth}
  \includegraphics[width=\linewidth]{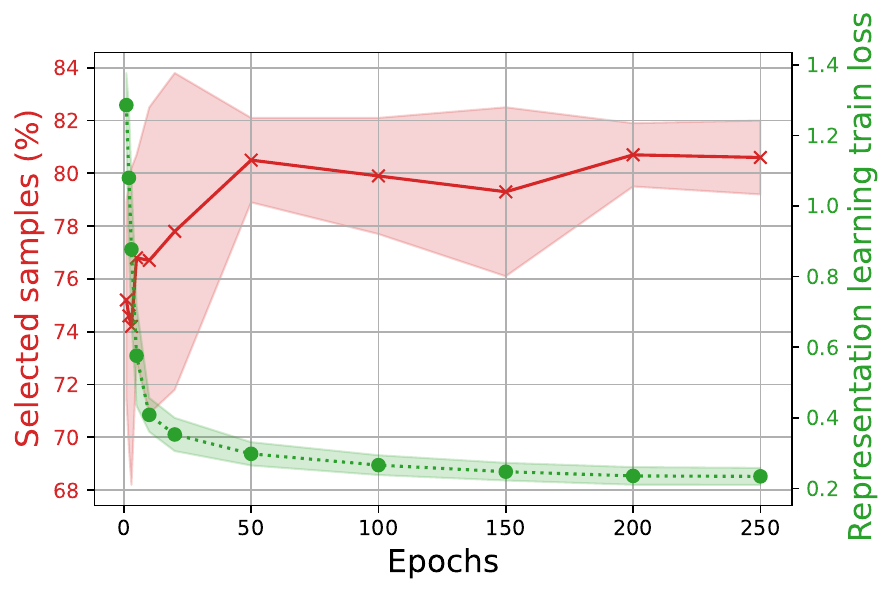}
\endminipage\hfill
\minipage{0.48\textwidth}
  \includegraphics[width=\linewidth]{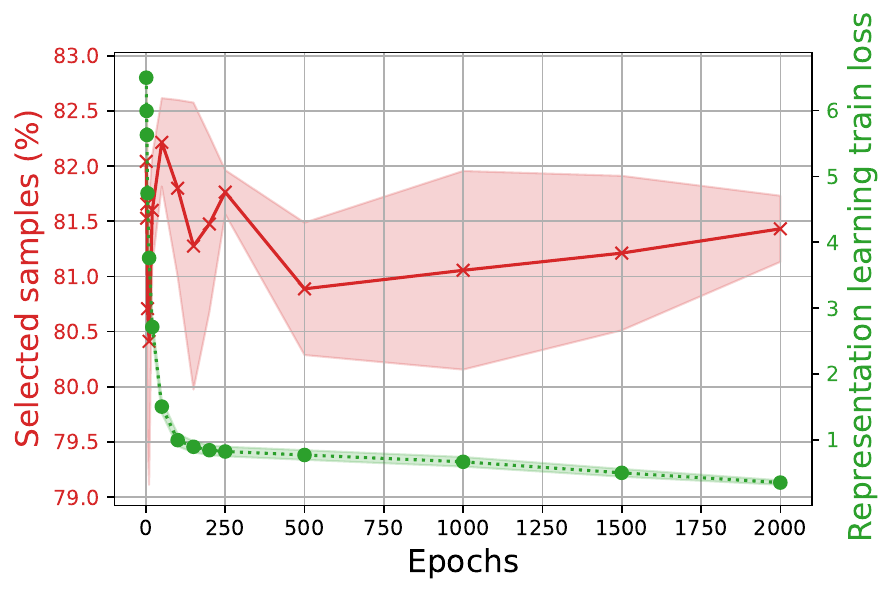}
\endminipage\hfill
\caption{Convergence graphs of {\algname} with (left) MVTec dataset, (right) CIFAR-10 dataset.}\label{fig:convergence_graph}
\end{figure}

Convergence is a well-known challenge in alternating training. It is more problematic in unsupervised learning settings due to the lack of validation set to evaluate the performance convergence. In the proposed framework, feature extractor loss is used for the convergence criteria (if no improvement is observed in the loss for 5 epochs) and data refinement block is intermittently updated with much lower frequency. The data refinement block consists of shallow one-class classifiers (not neural networks) trained on top of learned representations. Empirically, when the feature extractor's loss converges, the data refinement block is also converged.
Fig.~\ref{fig:convergence_graph} illustrate the convergence graphs of {\algname} with MVTec and CIFAR-10 datasets.

\section{Computational complexity}\label{sect:appendix_computational_complexity}
Note that when applying {\algname} on top of representation learning, the computational complexity of the representation learning part is not changed. With {\algname}, the additional computations come from training the ensemble models on top of learned representations. Note that we use shallow one-class classifiers (such as GDE or one-class SVM) for submodels; thus, the additional computational complexity is marginal. We would like to mention that all the experiments are done on a single V100 GPU and each experiment needs at most 12 hours for training (the additional training time caused by the {\algname} framework is an average 13.1\% of the total training time). The computations of the ensemble parts can be further improved by the model parallelization. 

\section{Implementation details for GOAD~\citep{bergman2020classification}}
\label{sec:app_goad}

A classification-based AD method, GOAD~\citep{bergman2020classification}, has demonstrated strong AD performance on tabular datasets. 
Unlike previous works~\citep{golan2018deep,hendrycks2018deep} that formulate a parametric classifier for multiple transformation classification, GOAD employs distance-based classification of multiple transformations. For the set of transformations $T_{m}:\mathcal{X}\,{\rightarrow}\,\mathcal{D}$, $m\,{=}\,1,...,M$, the loss function of GOAD is written as in Eq.~\ref{eq:goad_loss} with the probability defined in Eq.~ \ref{eq:goad_original}.
\begin{gather}
    \mathcal{L}=-\mathbb{E}_{m,x}\big[\log P(m|T_{m}(x))\big], \label{eq:goad_loss}\\
    P(\hat{m}|T_{m}(x)) = \frac{\exp(-\Vert f(T_{m}(x)) - c_{\hat{m}}\Vert^{2})}{\sum_{n}\exp(-\Vert f(T_{m}(x)) - c_{n}\Vert^{2})},\label{eq:goad_original}
\end{gather}
where the centers $c_{m}$'s are updated by the average feature over the training set. While it is shown to perform well~\citep{bergman2020classification}, we find that the distance-based formulation is not necessary, and we achieve the similar performance, if not worse, to \citep{bergman2020classification} using a parametric classifier when computing the probability:
\begin{equation}
    P(\hat{m}|T_{m}(x)) = \frac{\exp\left(w_{\hat{m}}^{\top}f(T_{m}(x)) + b_{\hat{m}}\right)}{\sum_{n}\exp\left(w_{n}^{\top}f(T_{m}(x)) + b_{n}\right)}\label{eq:goad_ours}
\end{equation}
The formulation in Eq.~\ref{eq:goad_ours} is easier to optimize than its original form in Eq.~\ref{eq:goad_original} as it can be fully optimized with backpropagation without alternating updates of feature extractor $f$ and centers $c_{m}$. 
Once we learn a representation by optimizing the loss in Eq.~\ref{eq:goad_loss} using Eq.~\ref{eq:goad_ours}, we follow a two-stage one-class classification framework of \citep{sohn2020learning} to construct a set of Gaussian density estimation OCCs for each transformation. Finally, we aggregate a maximum normality scores from a set of classifiers as the normality score.

In Table~\ref{tab:goad_implementation_details}, we summarize the implementation details, such as network architecture or hyperparameters, and AD performance under clean training data setting that reproduces the results in \citep{bergman2020classification}.

\begin{table}[h!]
    \caption{The AD performance under clean only data setting of GOAD in \citep{bergman2020classification} and our implementation. Our implementation demonstrates comparable, if not worse, performance to those reported in \citep{bergman2020classification}. Our implementation also shares most hyperparameters across datasets except the $M$, the number of transformations, and the train steps, which are closely related to the size of training data.}
    \centering
    \begin{tabular}{c|c|c|c|c}
        \toprule
        Datasets & KDD & KDD-Rev & Thyroid & Arrhythmia \\
        \midrule
        F-score~\citep{bergman2020classification} & 98.4{\scriptsize$\pm$0.2} & 98.9{\scriptsize$\pm$0.3} & 74.5{\scriptsize$\pm$1.1} & 52.0{\scriptsize$\pm$2.3} \\
        F-score (ours) & 98.0{\scriptsize$\pm$0.2} & 95.0{\scriptsize$\pm$0.2} & 75.1{\scriptsize$\pm$2.4} & 54.8{\scriptsize$\pm$3.2} \\
        \midrule
        $f$ (feature) & \multicolumn{4}{c}{$\big[$Linear(8), LeakyReLU(0.2)$\big]$ $\times$5}\\
        Optimizer & \multicolumn{4}{c}{Momentum SGD (momentum${=}\,0.9$)} \\
        Learning rate & \multicolumn{4}{c}{0.001} \\
        Batch size & \multicolumn{4}{c}{64$\times M$} \\
        \midrule
        L2 weight regularization & \multicolumn{4}{|c}{{0.00003}} \\
        \midrule
        Random projection dimension & \multicolumn{4}{|c}{{32}} \\
        \midrule
        $M$ & \multicolumn{2}{c|}{32} & \multicolumn{2}{c}{256} \\
        Train steps & \multicolumn{2}{c|}{$2^{10}$} & \multicolumn{2}{c}{$2^{16}$} \\
        \bottomrule
    \end{tabular}
    \label{tab:goad_implementation_details}
\end{table}

\end{document}